\documentclass[runningheads]{llncs}
\usepackage[T1]{fontenc}
\usepackage{graphicx}
\usepackage{cite}
\usepackage{mathtools}
\usepackage{xcolor}
\usepackage{booktabs}
\usepackage{adjustbox}
\usepackage{array}
\usepackage{tabularx}
\usepackage{hyperref}
\usepackage{manyfoot}
\usepackage{multirow}

\begin{document}
\title{VALE: A Multimodal Visual and Language Explanation Framework for Image Classifiers using eXplainable AI and Language Models}

\author{Purushothaman~Natarajan~$^{1}$ and Athira~Nambiar$^{1}$}%

\date{%
    $^1$SRMIST, Chennai, India \\
    [2ex]%
     \today}

\author{Purushothaman Natarajan\inst{1} \and
Athira Nambiar*\inst{1}}
\institute{Department of Computational Intelligence,\\
Faculty of Engineering and Technology,\\
SRM Institute of Science and Technology\\
Kattankulathur, Tamil Nadu, 603203, India\\ 
\email{c30945@srmist.edu.in, athiram@srmist.edu.in}}

\maketitle              
\vspace{-.8cm}
\begin{abstract}
Deep Neural Networks (DNNs) have revolutionised various fields, enabling the automation of tasks and minimizing human error. However, the internal workings of DNN and the rationale behind its decision-making processes remain unknown due to their `black-box' nature. As a result, the results' lack of interpretability has limited the use of these models in high-risk scenarios. In an effort to explain and interpret the internal workings of DNNs, a new field of research i.e. eXplainable Artificial Intelligence (XAI), is emerged. Nevertheless, in real-world scenarios, XAI encounters certain challenges, such as semantic gap in machine-human understanding,  trade-off between interpretability \& performance and context-specific explanation. To overcome such limitations, we propose a novel multimodal \textbf{V}isual \textbf{a}nd \textbf{L}anguage \textbf{E}xplanation framework named as (\textbf{``VALE''}) using explainable AI and language models. Upon the visual explanations provided by the XAI tool, an advanced zero-shot image segmentation model and a visual language model are incorportaed to extract the corresponding textual explanation. This multimodal visual \& textual explanation bridges the semantic gap between human and machine interpretation of the results, by providing human-compliant results. In this paper, we conduct a pilot study of the  VALE framework on image classification tasks. In particular, Shapley Additive Explanations (SHAP)  are applied to the classified images to identify the most influential regions. Further, the object of interest is obtained using the Segment Anything Model (SAM), and the corresponding explanation are achieved via the state-of-the-art pre-trained Vision Language Models (VLM). Extensive experimental studies are conducted on two datasets: the  ImageNet dataset and a tailor-made underwater SONAR image dataset, demonstrating real-world application in underwater image classification. Results show the promising performance of VALE multimodal explanation framework.

\vspace{-0.3cm}

\end{abstract}

\vspace{-.7cm}
\section{Introduction}
\label{Sec: Introduction}
\vspace{-0.35cm}
Image classification is a task that involves predicting class
labels to images, based on their visual content~\cite{rawat2017deep}. It is extensively used in various applications, such as medical imaging, object detection and autonomous driving. The realm of image classification has experienced significant advancement over time, with the integration of deep learning methodologies. However, these models remain opaque, making them `black-boxes' by nature i.e. they fail to explain their own decisions in a human-compliant manner.

To overcome this challenge, a set of techniques and methods, which are aimed at enhancing the interpretability of AI systems' decisions, known as "Explainable Artificial Intelligence (XAI)" has been developed. Various XAI tools such as Local Interpretable Model Agnostic Explanation (LIME)~\cite{ribeiro2016should}, Shapley Additive exPlanation (SHAP)~\cite{lundberg2017unified}, Class Activation Mapping (CAM)~\cite{selvaraju2017grad}, and Layer-wise Relevance Propagation (LRP)~\cite{bach2015pixel}, have been recently introduced in this field. All of these methods draw attention to highlight crucial elements within the region of interest and provide visual masks to offer a visual explanation to the predictions made by the image classification model. Nevertheless, all the aforementioned XAI approaches still lack in explaining the result in a natural human comprehendable way i.e. textual explanation. This creates a semantic gap in the human-machine way of interpreting the result and demands a sufficient understanding of the explainer to interpret the predictions made by the deep neural network~\cite{samek2017explainable}. Also, many XAI-based visual explainers are suboptimal to provide context-aware explanations as well.

In this work, we propose a novel multimodal \textbf{Visual and Language Explanation} framework (\textbf{VALE}), that provides not only the visual explanation of the image classifier, but also its textual counterpart. This dual explanation is facilitated via a visual explanation from XAI tool and its textual explanation  generated with the help of zero-shot image segmentation models and pre-trained vision Language Models (VLMs). This combination bridges the semantic gap that exists between the two modalities of the explainers. In particular, the image classification results are explained visually through the SHAP explainer, a post-hoc model agnostic technique that makes use of Shapley scores to identify the most influential regions in the image.
Further, a bench-marking segmentation model, i.e., the Segment Anything Model (SAM)~\cite{kirillov2023segment} is used to segment the object based on the top-most influential region i.e area that is of interest for the predicted label, thereby offering a second visual explanation in a straightforward and tangible way. This segmented region is further described using a VLM, that provides human-complaint textual explanation from the visual counterpart, with the help of domain-specific language instruction/ prompt.

To showcase the efficacy of the proposed VALE architecture, it is experimented on a generic image classification task using the ImageNet dataset in our pilot study. Further, we also apply for a specific case study of underwater SONAR image classification, upon a custom-built image classification model ~\cite{chungath2023transfer}. In this investigation, we also depict how the VALE model can be fine-tuned for `in-the-wild' applications with the help of transfer learning and specialized prompt engineering. The key contributions of the paper are summarized as follows:
\vspace{-0.3cm}
\begin{itemize}
    \item Proposal of a novel multimodal XAI framework \textbf{``VALE: Visual and Language Explanation"} for Image Classification task.
    \item Integration of pretrained segmentation and image-captioning VLM models to augment eXplainable AI from visual explainer to textual explainer realm.
    \item Prompt engineering to optimize the VLM response and performance analysis of the proposed framework  for real-world application in underwater SONAR image classification.
\end{itemize}
\vspace{-0.2cm}
The rest of the paper are organized as follows: the related works on XAI for image classification and image-to-text explainer is presented in Section~\ref{Sec: Related work}. The overall pipeline of the proposed  VALE architecture is explained in Section~\ref{Sec: Methodology}. Further, the experimental setup and experimental results are summarized in Section~\ref{Sec: Experimental Setup} and Section~\ref{Sec: Experimental Results}, respectively. Finally, the conclusion and future works are summarized in Section~\ref{Sec: Conclusion}.

\vspace{-0.6cm}
\section{Background and Related Work}
\label{Sec: Related work}
\vspace{-0.25cm}
\subsection{Explainable AI(XAI) for Image Classification}
\label{Subsec: XAI for Image classification}
\vspace{-0.25cm}
The XAI techniques are classified into two categories: model-specific and model-agnostic. Model-specific techniques are designed to analyze and explain the behaviour of a specific ML model, taking into account its unique architecture and complexities. These techniques are particularly useful for models such as decision trees and logistic regression~\cite{samek2017explainable}. On the other hand, model-agnostic techniques aim to provide explanations that are independent of the particular model being used. Model-agnostic techniques such as LIME, SHAP, CAM, etc. are valuable for explaining DNN models. LIME, developed by Ribeiro et al.n~\cite{ribeiro2016should}, explains image classification model predictions by locally perturbing the input sample and fitting it in a linear model to find the pertinent features for the prediction. Similarly, Bach et al.~\cite{bach2015pixel} plot the pixel-wise contribution in each layer of the neural network to explain the model decision on a heat map. In 2017, Lundberg et al~\cite{lundberg2017unified}. introduced an open-source library to use Shapley scores from cooperative game theory to explain black-box model predictions on structured and unstructured data. SHAP explains classification model predictions in various domains, including medical~\cite{samek2017explainable}, agriculture~\cite{abdollahi2021urban}, aerial imagery~\cite{ayush2020generating}, etc. Then, the highly popular XAI technique for image and video-based DL models by Selvaraju et al.~\cite{selvaraju2017grad} utilizes Gradient-weighted Class Activation Mapping (Grad-CAM) to visualize DNN predictions as heat maps. In a recent study, Sun et al.~\cite{sun2024explain} used a combination SHAP and LIME explainer to segment the object of interest in the image using the pre-trained model SAM to provide a better visual explanation over using a heat map provided by SHAP explainer.

\vspace{-0.45cm}
\subsection{Image-to-text Explainer}
\label{Subsec: Image-to-text explainer}
\vspace{-0.25cm}
Image captioning, or image-to-text explanation, is the process of creating textual descriptions for images~\cite{wang2022git}. These captions serve as a textual representation of the visual content contained within an image. Wang et al.~\cite{wang2022git} employed an image encoder and text decoder combination to produce captions for images. Most captioning models have a similar architecture, and visual-textual explanations are rarely studied together. However, there are models that explain the process of converting images to text. In a study by Dewi et al.~\cite{dewi2023xai}, SHAP was used to analyze the performance of Azure Cognitive Service's image captioning model and other publicly available models in generating captions. Sahay et al.~\cite{sahay2021approach} utilized LIME to visualize the image portion associated with a caption word. Han et al.~\cite{han2020explainable} employ an attention mechanism to map objects using a Mask Region-based Convolutional Neural Network (Mask-RCNN) and generate textual descriptions. The Greybox AI~\cite{bennetot2022greybox} authors mapped predictions and explanations using a latent space predictor and explainable latent space to offer a superior explanation compared to the other papers.

In contrast to the aforementioned approaches, which provide either visual or textual explanations, none of the models utilize XAI for multimodel explanation, for the first time, to the best of our knowledge, VALE offers a multimodal explanation, i.e., both visual and textual explanations, in a human-compliant manner via an explainer, in which end users do not require domain expertise or understanding of the underlying explainer to comprehend the explanation.

\vspace{-.55cm}
\section{Methodology:  Visual and Language Explanation}
\label{Sec: Methodology}
\begin{figure}[h!]
    \vspace{-1cm}
    \centering
    \includegraphics[width=\textwidth]{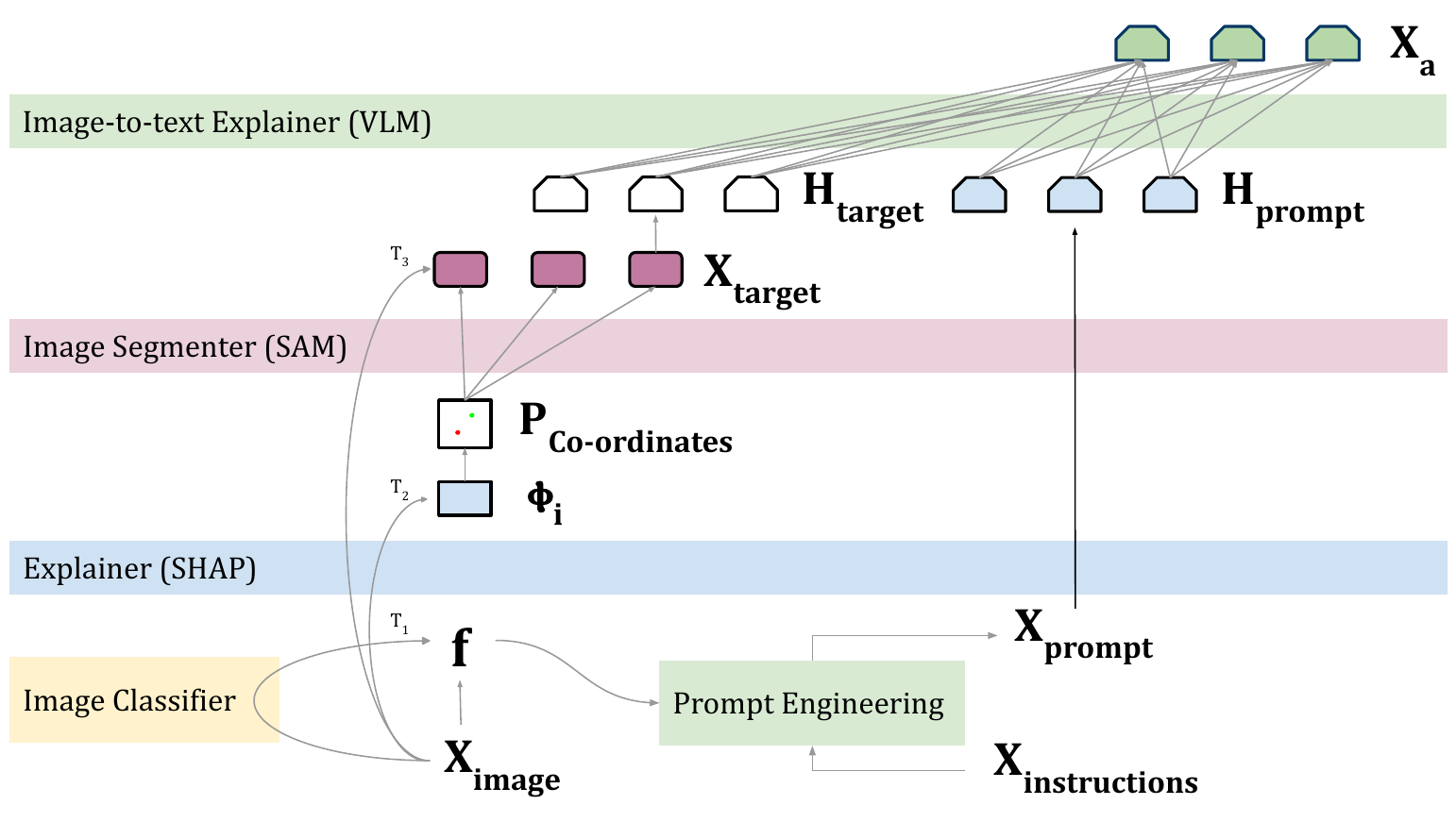}
    \vspace{-0.7cm}
    \caption{Architecture of VALE: Visual and Language Explainer framework.}
    \label{fig:Overall Architecture}
    \vspace{-0.65cm}
\end{figure}

\vspace{-0.25cm}
In this section, the multimodal Visual and Language Explanation (VALE) framework for image classification task is explained. Referring to Fig.~\ref{fig:Overall Architecture}, VALE consists of four separate components: Image classifier, Explainer (SHAP), Image segmenter (Segment Anything model) and Image-to-Text explainer (VLM). All of these modules are detailed in the forthcoming sections.

\vspace{-0.65cm}
\subsection{Image Classifier}
\label{Subsec: methodology - image classification}
\vspace{-0.25cm}
Image classification is a technique for classifying images by utilizing historical data (training data). Convolutional Neural Networks (CNNs) are a type of ANN specifically used in the field of pattern recognition within images and are more efficient for image classification compared to traditional machine learning models~\cite{o2015introduction}. The prediction using CNNs is represented by the following expression.

\vspace{-0.25cm}
\begin{equation}
    f = \text{softmax}\left(W^{[L]} \cdot \text{flatten}\left(\text{pooling}\left(g\left(W^{[l]} * X_{\text{image}} + b^{[l]}\right)\right)\right) + b^{[L]}\right).
\end{equation}
\vspace{-0.35cm}

where \(X_{\text{image}}\) is the input image, \(W^{[l]}\) and \(b^{[l]}\) are the weights and biases of layer \(l\), \(g\) is the activation function (typically ReLU), \(*\) represents convolution, \(\text{pooling}\) represents pooling operations like max pooling, \(\text{flatten}\) converts the pooling layer output into a vector, \(W^{[L]}\) and \(b^{[L]}\) are the weights and biases of the output layer, \(\text{softmax}\) is the activation function used for multiclass classification and \(f\) is the predicted label. The process of developing and fine-tuning these architectures requires a significant amount of time and computation. Therefore, the existing architectures and publicly accessible pre-trained models, are utilized as a starting point to predict the class label of the input image. For custom datasets and curated models, transfer learning (i.e. knowledge gained through one task or dataset is used to improve model performance on another related task and/or different dataset) is leveraged ~\cite{chungath2023transfer}.

\vspace{-.55cm}
\subsection{Explainer: SHapley Additive exPlanations (SHAP)}
\label{Subsec: methodology - SHAP}
\vspace{-.25cm}
The SHAP explainer offers a global approach to explain the predictions made by a black-box model. This explainer is based on cooperative gaming theory and the concept of Shapley values~\cite{winter2002shapley},~\cite{lundberg2017unified}. The model's prediction is determined by computing the Shapley scores for each feature. The following expression computes the score of feature \(i\) on the overall prediction.

\vspace{-0.2cm}
\begin{equation}
    \phi_i = \sum_{S \subseteq F \setminus \{i\}} \frac{|S|! (|F| - |S| - 1)!}{|F|!} \left[ f_{S \cup \{i\}}(x_{S \cup \{i\}}) - f_S(x_S) \right].
    \vspace{-0.1cm}
\end{equation}

where, \(\phi_i\) is the contribution (SHAP value) of feature \(i\), \(F\) is the set of all features, \(S\) is a subset of \(F\) excluding the feature \(i\), \(|S|!\) is the factorial of the size of the set \(S\), representing the number of ways to arrange \(S\), \(|F| - |S| - 1\) is the number of features not in \(S\) excluding \(i\), and its factorial represents the number of ways to arrange the remaining players, \(|F|!\) is the factorial of the total number of features, representing the total number of ways to arrange all features, \(f_{S \cup \{i\}}(x_{S \cup \{i\}})\) is the value function for the coalition \(S\) including feature \(i\), \(f_S(x_S)\) is the value function for the coalition \(S\) without feature \(i\). The above equation can also be written with respect to the prediction from model \(f\) to the specific input \(x\),

\vspace{-.4cm}
\begin{equation}
    \phi_i(f, X_{\text{image}}) = \sum_{\mathclap{z' \subseteq x'}} \frac{|z'|! (M - |z'| - 1)!}{M!} \left[ f_x(z') - f_x(z' \setminus \{i\}) \right].
    \vspace{-0.15cm}
\end{equation}

where, \(|z'|\) is the number of non-zero entries in \(z'\), and \(z' \subseteq x' \) represents all \(z’\) vectors where the non-zero entries are a subset of the non-zero entries in \(x'\). Based on these scores \(\phi\) for each feature, a heatmap is overlaid on the image to indicate the most important and least important features for the predicted class. Further, 

\vspace{-.5cm}
\begin{equation}
    P_{\text{co-ordinates}} = \arg\max_{1 \leq i \leq n} \left( \phi_i(f, X_{\text{image}}) \right).
    \vspace{-0.05cm}
\end{equation}

where, \(n\) is the total number of SHAP values computed for the input image \(X\), \(\phi(f, X_{\text{image}})\) represents the calculated SHAP values and \(P_{\text{coordinates}}\) refers to the index \(i\) that corresponds to the highest value of \(\phi_i(f, X_{\text{image}})\) among all \(\phi_i(f, X_{\text{image}})\) values for \(i = 1, 2, ..., n\). In our case, the index is the coordinates from the input image with the highest SHAP value.

\vspace{-.65cm}
\subsection{Image Segmenter: Segment Anything Model}
\label{Subsec: methodology - image segmentation}
\vspace{-.25cm}
Image segmentation is the technique of partitioning an image into distinct groups of pixels, known as image segments. This process aids in object identification and creates boundaries within the image based on areas of interest, resulting in a more meaningful and simplified analysis. We employ the instance segmentation model viz. Segment Anything Model (SAM)~\cite{kirillov2023segment} as the de-facto model to segment the region of interest (target object) from the image. SAM is chosen due to its robust zero-shot performance and its ability to generate segmentation using prompts such as points, boxes, and text. Refering to Section \ref{Subsec: methodology - SHAP}, SHAP explainer identifies and highlights the specific regions in the image that have the highest and lowest contribution to the predicted class. The coordinates \(P_{\text{coordinates}}\) with the highest SHAP score in the image are used as the prompt  (point) to generate the zero-shot image segmentation. This segments the input image and extracts the target object \(X_\text{target}\) from the entire image.

\vspace{-.55cm}
\subsection{Image-to-text Explainer via Language Model \& Prompt Engineering}
\label{Subsec: methodology - image-to-text explanation}
\vspace{-0.25cm}

Image-to-text explanation, or image captioning, refers to the process of generating textual descriptions or textual depictions of the visual content present in an image. Most image captioning models typically follow an encoder-decoder architecture~\cite{liu2024visual}. The encoder is usually a Convolutional Neural Network (CNN) that captures relevant information from the image. The decoder, on the other hand, is a Recurrent Neural Network (RNN) that decodes the captured visual information into a descriptive sequence of text~\cite{xu2015show}. The key component in such models is the attention mechanism, which allows them to focus on the relevant parts of the image while generating each word in the caption. This captioning can be extended to visual question answering, wherein the user can interact with the generated captions and the user can also provide hints about the image to the model with prompts to get a highly relevant response from the VLM.

Prompt engineering refers to the systematic approach of designing and improving the input queries (prompts) to obtain the desired response from the Language Model (LM). It expands the functionalities of language models without altering the core parameters, and it improves and directs language models to produce the desired output. For a custom model, it is important to fine-tune and refine the prompt to achieve the desired output, especially in specialized domains where the input data (image) is collected using non-standard processes. In such cases, mentioning the specific technique used in the input prompt leads to better output compared to the standard output. In our case, we strengthen the instructions by incorporating the predicted label from the classification model into the language instructions \(X_\text{instructions}\) to develop the prompt \(X_\text{prompt}\).

Developing such large Vision Language Models (VLMs) is time-consuming and computationally expensive. Therefore, our study leverages the advantage of existing pre-trained VLMs to generate captions for the segmented image from SAM, \(X_\text{target}\). Referring to Fig.~\ref{fig:Overall Architecture}, SAM predicts three masks, each with different confidence scores; the image with the highest confidence score is processed using a trained language model, which converts the image into a sequence of visual tokens \(H_\text{target}\). The \(X_\text{prompt}\) is processed using the same language model, which converts the text into a sequence of textual tokens \(H_\text{prompt}\). The language model processes the prompt and the image tokens together as a conditional vector to provide a textual description \(X_\text{a}\).

We employ VLMs integrated with a vision tower and a language decoder to provide a textual explanation. This model works with two inputs, specifically the prompt (instruction) and the image, to generate a textual description. The equation for generating the textual explanation $X_a$ given an image $X_v$ and prompt $X_{\text{prompt}}$ can be represented as follows:

\vspace{-.65cm}
\begin{equation}
    p(X_a|X_{\text{target}}, X_{\text{prompt}}) = \prod_{i=1}^{L} p_{\theta}(x_i|X_\text{target}, X_{\text{prompt}}, <i, X_a, <i).
\end{equation}
\vspace{-.4cm}

where, \(p(X_a|X_{\text{target}}, X_{\text{prompt}})\) represents the likelihood of target answers \(X_a\) given the image \(X_{\text{target}}\) and instruction \(X_{\text{prompt}}\). \(p_{\theta}(x_i|X_v, X_{\text{prompt}}, <i, X_a, <i)\) represents the conditional probability of token \(x_i\) based on image, prompt, previous tokens, and answer tokens. The sequence length is \(L\). This approach serves as an initial step in creating personalized image captioning for a custom dataset. However, we employ it to produce a detailed rationale for the prediction made by the classifier by employing a personalized prompt \(X_{\text{prompt}}\) and a segmented image with the higher confidence score \(X_{\text{target}}\).

\vspace{-.6cm}
\section{Experimental Setup}
\label{Sec: Experimental Setup}
\vspace{-.35cm}
In this section, the dataset used for training, the implementation details, and the evaluation metrics used to assess the model performance are described.

\vspace{-.5cm}
\subsection{Dataset for Learning}
\label{Subsec: Dataset}
\vspace{-.15cm}
\subsubsection{ImageNet Dataset:}
\label{dataset: ImageNet}
ImageNet is an open-source dataset consisting of 15 million labeled images and 1000 distinct labels~\cite{deng2009imagenet}. Each label has a minimum of 1000 images associated with it and is one of the most widely used datasets for training image classification models. One of the main reasons for choosing this dataset is the availability of a diverse range of pre-trained models, which can be used for both commercial and research purposes~\cite{deng2009imagenet}. The images in this dataset were obtained from numerous sources and have varying dimensions. 

\vspace{-.6cm}
\subsubsection{SONAR Dataset:}
\label{dataset: sonar}
The availability of datasets for critical domains such as defence and medicine is limited due to their sparse and confidential nature. Therefore, we showcase the efficacy of the proposed architecture in the field of underwater SONAR imagery, using a curated tailor-made dataset. This SONAR data is collected by several publicly available datasets that are published for academic research i.e. \textit{Seabed Objects KLSG}~\cite{huo2020underwater} and \textit{Sonar Common Target Detection Dataset (SCTD)}~\cite{zhang2021self}. In this dataset, we have obtained 753 images of ships, 123 images of planes, and 578 images of the seafloor.


\vspace{-.5cm}
\subsection{Evaluation Metrics}
\label{Subsec: Evaluation metrics}
\vspace{-.25cm}

The performance of the classification models is assessed using standard evaluation metrics i.e. Accuracy, Precision, Recall and F1-score~\cite{vujovic2021classification}. There are no established methodologies for quantifying the performance of the SHAP explainer. However, the performance of the explainer can be visually assessed by analyzing the distribution of scores in the image through a heatmap. The relevant feature in the image should receive a high SHAP value, while non-relevant features should receive a low value. The performance of the Segmentation model can be assessed using Intersection over Union (IoU)~\cite{divvala2009empirical}. However, since SAM is pre-trained, its performance is not evaluated in our study, instead the confidence score from SAM's prediction are utilized. The Image captioning models can be accessed using BLEU (Bilingual Evaluation Understudy)~\cite{papineni2002bleu}. The efficacy of the proposed framework in delivering textual explanations is accessed with BLEU scores through manually annotated samples (Refer Table~\ref{tab:reference for ImageNet} and Table~\ref{tab:sonar_image_descriptions}). Note that the evaluation is limited to a small number of human annotated samples due to a lack of annotated data and computation for both the ImageNet and SONAR datasets.

\vspace{-.6cm}
\subsection{Implementation Details}
\label{Subsec: Implementation details}
\vspace{-.35cm}
In this study, five prominent pre-trained models, namely VGG16, Xception, InceptionV, ResNet50, and DenseNet121 are used as the image classifier models on the ImageNet dataset. To maintain consistency, we adopt 224 * 224 image dimensions as utilized in the pre-trained models. For the SONAR counterpart, we utilize transfer learning to develop an image classification model by employing DenseNet121, customized with two active layers consisting of 1024 and 512 neurons, respectively. Additionally, we incorporate a dropout layer with a rate of 0.25 and a batch normalization layer. We train the model using an Adam optimizer with a learning rate of 0.0001 and a batch size of 16. For the SHAP explainer, we select a batch size of 50 and specified the maximum evaluation parameter count to be 1000. We choose the zero-shot image segmentation model SAM (Segment Anything Model) for image segmentation. To segment the target object, we utilize the coordinates obtained from the SHAP explainer as the input prompt for SAM. We utilize pre-trained VLM's such as Large Language and Vision Assistant (LLaVA)~~\cite{liu2024visual}, Instuctblip~\cite{dai2023instructblip}, Generative Image-to-text Transformer (GIT)~\cite{wang2022git}, MiniCPM~\cite{hu2024minicpm} and InternLM~\cite{dong2024internlm}, with default parameters such as a temperature value of 0.2, no specified top P value, and a maximum output token limit of 1024, to provide textual explanations. Additionally, the prompt is engineered to align with our specific situation. The implementation is conducted on Google Colab, utilizing an A100 GPU with an allocation of 15GB for training and employing pytorch framework.

\vspace{-.655cm}
\section{Experimental Results}
\label{Sec: Experimental Results}

\vspace{-.45cm}
\subsection{ Experimental Analysis on the ImageNet dataset}
\label{Subsec: results - ImageNet analysis}
\vspace{-0.35cm}
The efficiency of the proposed architecture is accessed through random samples obtained from the ImageNet dataset and the results are explained below:

\vspace{-.65cm}
\subsubsection{Image Classifier:}
\label{Subsubsec: results - ImageNet image classification}
The image classifiers are pre-trained, hence they do not require any additional training. The accuracies of the models are summarized in Table~\ref{tab:model_comparison}. To accommodate computational constraints, the model with the smallest number of parameters and the smallest size i.e. DenseNet121, which achieves an accuracy of 92.3\% with 8.1 million parameters, is selected as the de-facto backbone network for further study. The input images are pre-processed and then directly predicted using the pre-trained model. For instance, for the image sample shown in Fig.~\ref{fig:SHAP explainer output - bald eagle}(a), the model predicts the image class as a `bald eagle' with a probability of 100\%. This prediction is further explained through the SHAP explainer.

\begin{table}[htbp]
    \vspace{-0.35cm}
    \centering
    \caption{Pre-trained Model Comparison.}
    \label{tab:model_comparison}
    \vspace{-0.3cm}
    \begin{adjustbox}{max width=1.3\textwidth}
        \begin{tabular}{@{}llllll@{}}
            \toprule
            Model       & Size (MB) & Top-1 Accuracy & Top-5 Accuracy & Parameters & Depth \\ \midrule
            VGG16       & 528       & 71.3\%         & 90.1\%         & 138.4M     & 16    \\
            ResNet50    & 98        & 74.9\%         & 92.1\%         & 25.6M      & 107   \\
            InceptionV3 & 92        & 77.9\%         & 93.7\%         & 23.9M      & 189   \\
            Xception    & 88        & 79.0\%         & 94.5\%         & 22.9M      & 81    \\
            DenseNet121 & 33        & 75.0\%         & 92.3\%         & 8.1M       & 242   \\ 
            \bottomrule
            \bottomrule
        \end{tabular}
    \end{adjustbox}
    \vspace{-0.8cm}
\end{table}

\vspace{-.55cm}
\subsubsection{SHAP Explainer:}
\label{Subsubsec: results - ImageNet SHAP explainer}
\begin{figure}[h!]
    \footnotesize
    \vspace{-0.7cm} 
    \centering
    {\fontsize{6}{8}\selectfont
    \begin{tabular}{@{}cccc@{}} 
        \begin{minipage}{0.24\textwidth}
            \centering
            \resizebox{\linewidth}{3cm}{\includegraphics{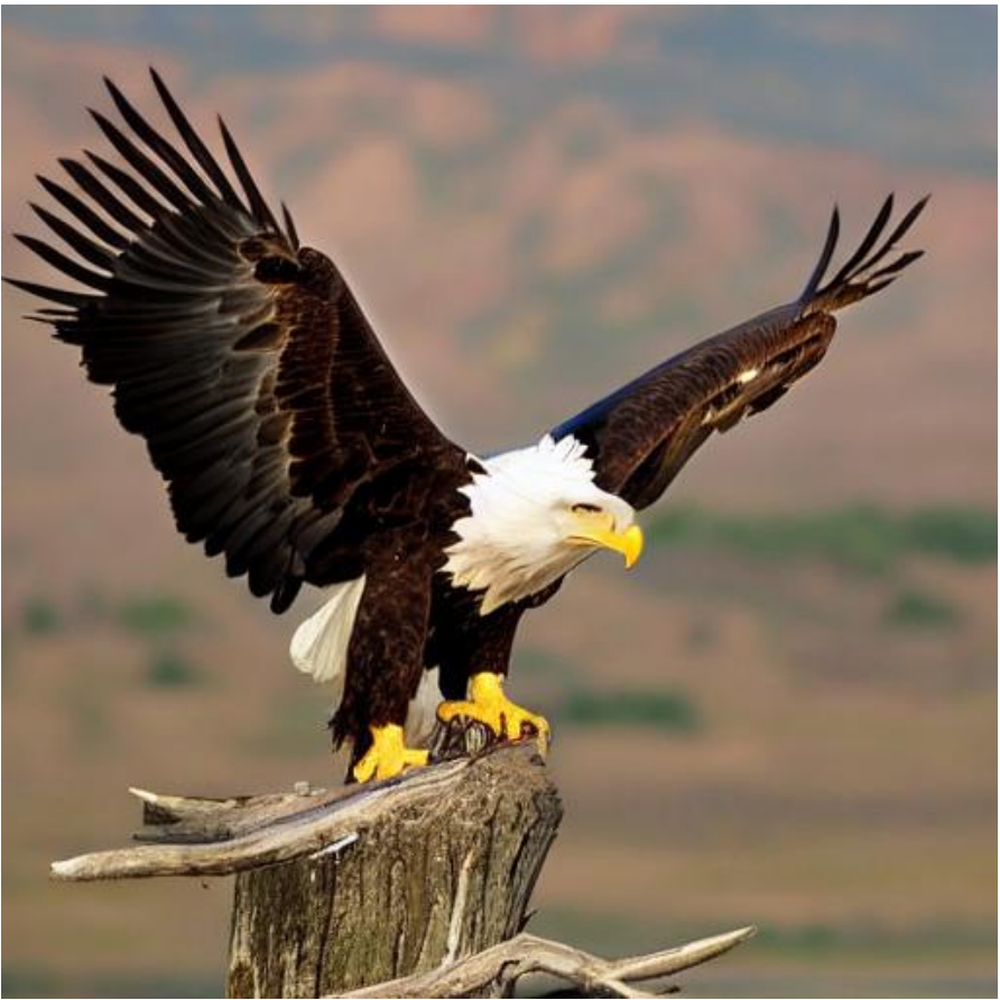}} 
            \par\vspace{0.5ex} 
            \textbf{(a) Input Image}
            \label{fig:sub: Original bald eagle input image}
        \end{minipage} &
        \begin{minipage}{0.24\textwidth}
            \centering
            \resizebox{\linewidth}{3cm}{\includegraphics{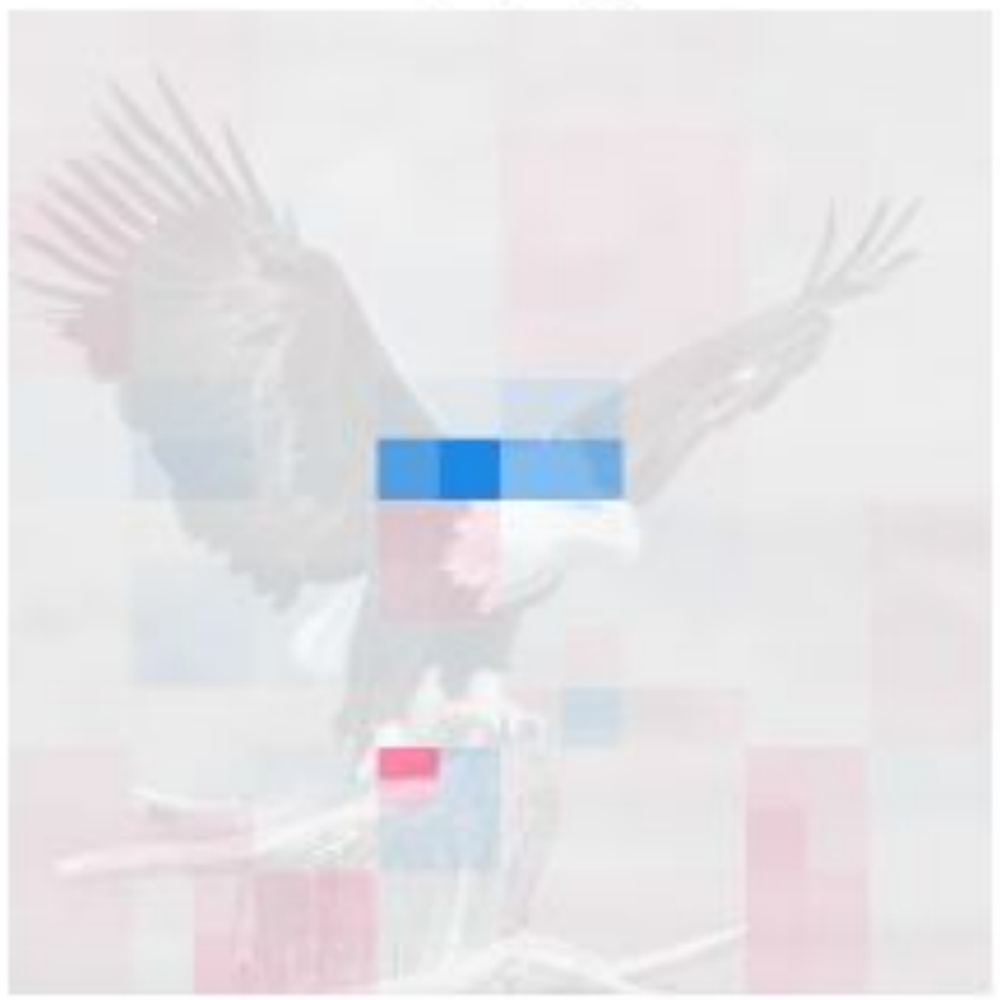}} 
            \par\vspace{0.5ex} 
            \textbf{(b) Explanation}
            \label{fig:sub: shap explainer output bald eagle}
        \end{minipage} &
        \begin{minipage}{0.24\textwidth}
            \centering
            \resizebox{\linewidth}{3cm}{\includegraphics{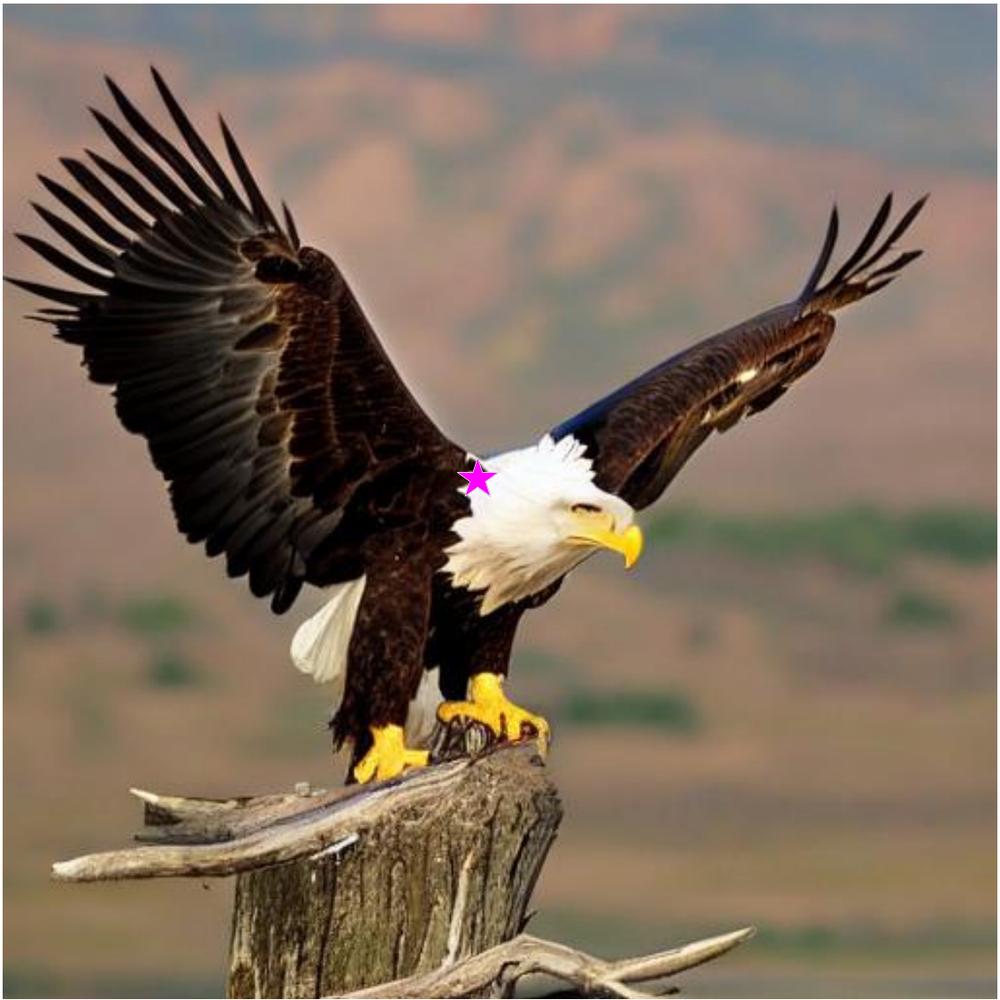}} 
            \par\vspace{0.5ex} 
            \textbf{(c) ROI}
            \label{fig:sub: ROI for bald eagle}
        \end{minipage} &
        \begin{minipage}{0.24\textwidth}
            \centering
            \resizebox{\linewidth}{3cm}{\includegraphics{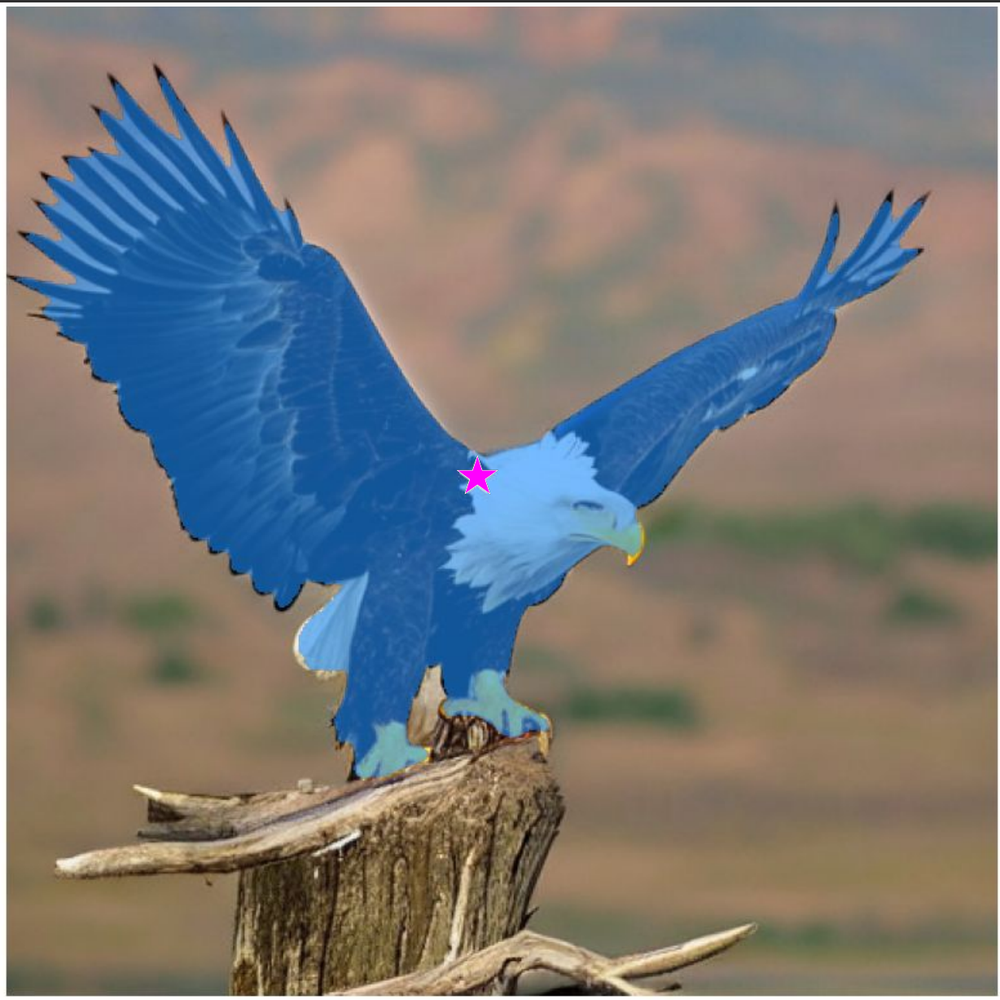}} 
            \par\vspace{0.5ex} 
            \textbf{(d) Generated Mask}
            \label{fig:sub: Mask 3 bald eagle}
        \end{minipage}
    \end{tabular}
    }
    \vspace{-0.2cm} 
    \caption{Output from the SHAP explainer (Explanation), the coordinate with the highest SHAP value (ROI - represented with a magenta star), and the generated mask for Bald Eagle.}
    \label{fig:SHAP explainer output - bald eagle}
    \vspace{-0.8cm} 
\end{figure}

The SHAP explainer has two parameters: the maximum evaluation parameter, which determines the total number of ways to arrange all features and the batch size. With a batch size of 50 and the maximum evaluation parameter count of 1000, the SHAP result for the bald eagle image is depicted in Fig. ~\ref{fig:SHAP explainer output - bald eagle}(b). From the SHAP values, the coordinates (\(P_{\text{coordinates}}\)) with the highest SHAP values are obtained, which are represented with a magenta star (Region of Interest (ROI)) in Fig.~\ref{fig:SHAP explainer output - bald eagle}(d).

\footnotetext{More results at available at the Supplementary link below: \url{https://drive.google.com/file/d/1Cli1hky2E-6pabmpBw_dZRFHbPesYG7W/}}

\vspace{-0.45cm}
\subsubsection{Segment the Object of Interest using SHAP Values:}
\label{Subsubsec: results - ImageNet image segmentation}
The coordinate with the highest SHAP value (\(P_{\text{coordinates}}\)) is provided as input to the zero-shot image segmentation model SAM. SAM generates distinct masks with varying confidence scores, where the mask with the highest score indicates the segmentation of highly similar regions or the entire object of interest. The SAM provides three masks for the bald eagle based on the given coordinate, as shown in Fig.~\ref{fig:SHAP explainer output - bald eagle}(d). The segmented mask in Fig.~\ref{fig:SHAP explainer output - bald eagle}(d) has a confidence score of 93.2\% accurate, as determined by SAM. The acquired image \(X_v\) is further explained using a VLM.

\vspace{-0.55cm}
\subsubsection{Ablation Study- Hyper-parameters of the SHAP Explainer:}
\label{Subsubsec: results - ImageNet SHAP hyper-parameters}

\begin{table}[h!]
    \centering
    \vspace{-.35cm}
    \caption{Effect of hyper-parameters of the SHAP explainer. Top row shows the number of maximum evaluation parameters and second row shows SHAP explanation. The last row shows the coordinate with the highest SHAP value (represented with a magenta star) and the corresponding segmented object mask obtained using SAM.}
    \vspace{-0.3cm}
    \begin{tabular}{|c|c|c|c|c|}
    \hline
    \multicolumn{5}{|c|}{Number of Maximum Evaluation Parameters} \\
    \hline
    100 & 200 & 300 & 500 & 1000 \\
    \hline
    \includegraphics[width=0.185\linewidth, trim=6.7cm 3.3cm 0.2cm 0.9cm, clip]{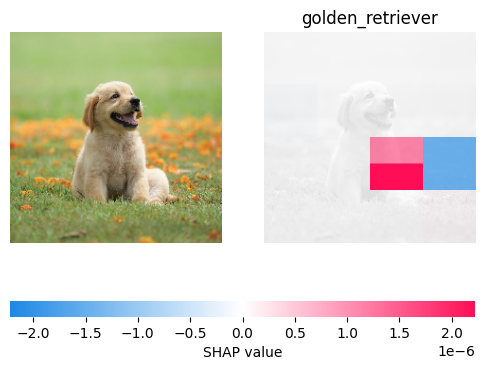} & 
    \includegraphics[width=0.185\linewidth, trim=6.7cm 3.3cm 0.2cm 0.9cm, clip]{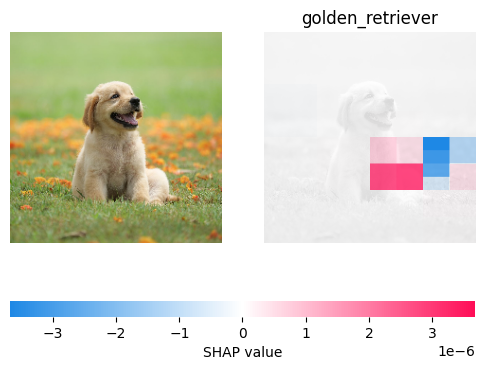} & 
    \includegraphics[width=0.185\linewidth, trim=6.7cm 3.3cm 0.2cm 0.9cm, clip]{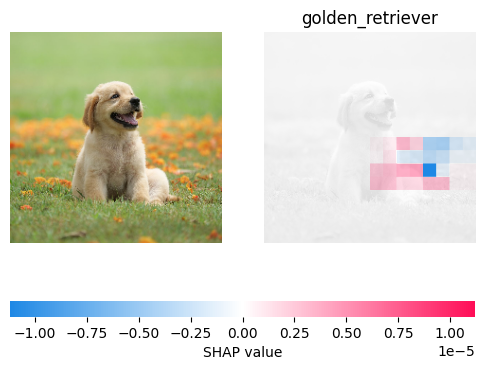} & 
    \includegraphics[width=0.185\linewidth, trim=6.8cm 3.3cm 0.2cm 0.9cm, clip]{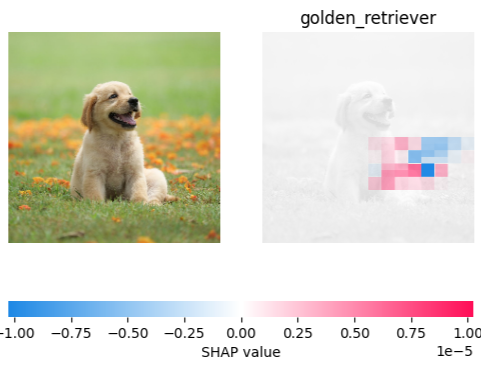} &
    \includegraphics[width=0.185\linewidth, trim=9.1cm 4.4cm 0.1cm 1cm, clip]{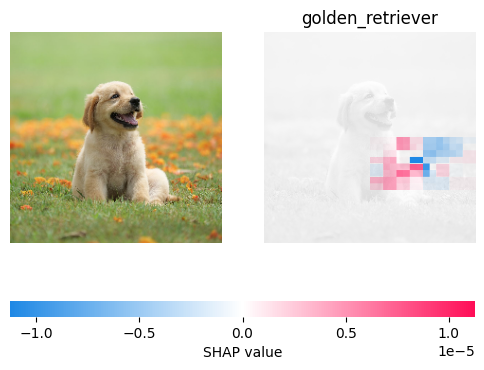} \\
    \hline
    \includegraphics[width=0.185\linewidth, trim=0.1cm 0.1cm 0.1cm 0.55cm, clip]{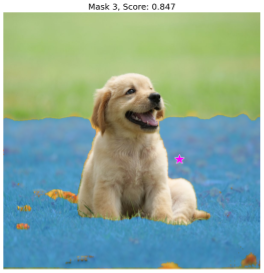} & 
    \includegraphics[width=0.185\linewidth, trim=0.1cm 0.1cm 0.1cm 0.7cm, clip]{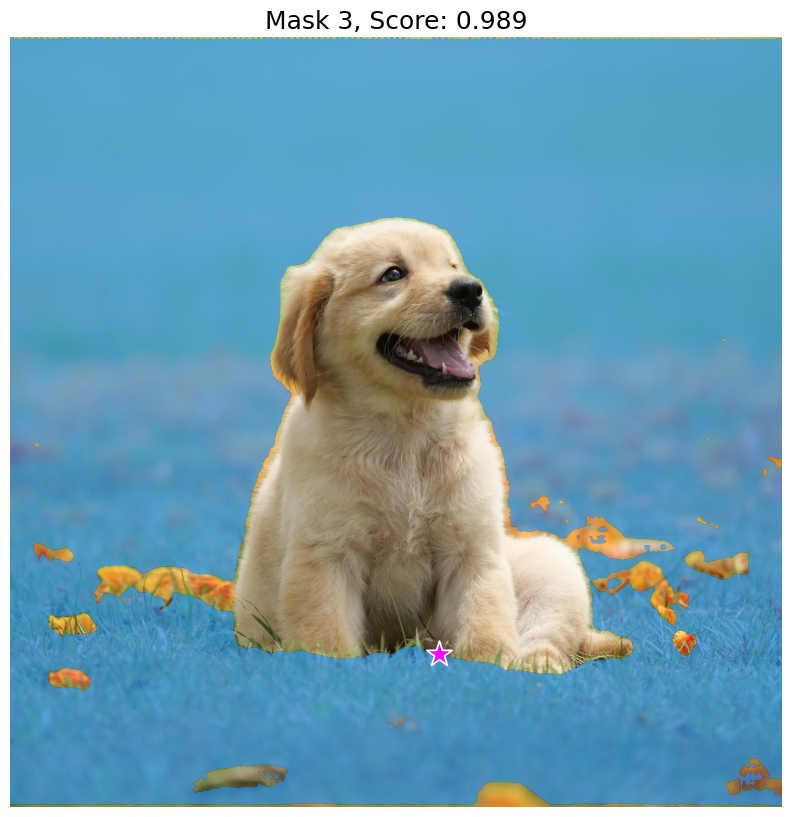} & 
    \includegraphics[width=0.185\linewidth, trim=0.1cm 0.1cm 0.1cm 0.7cm, clip]{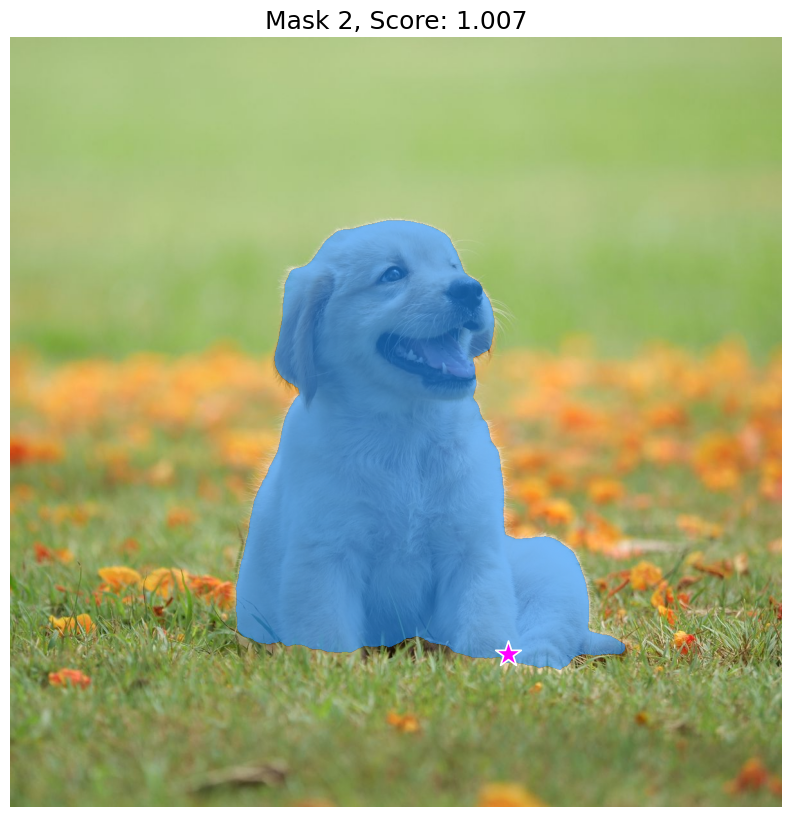} & 
    \includegraphics[width=0.185\linewidth, trim=0.1cm 0.1cm 0.1cm 0.7cm, clip]{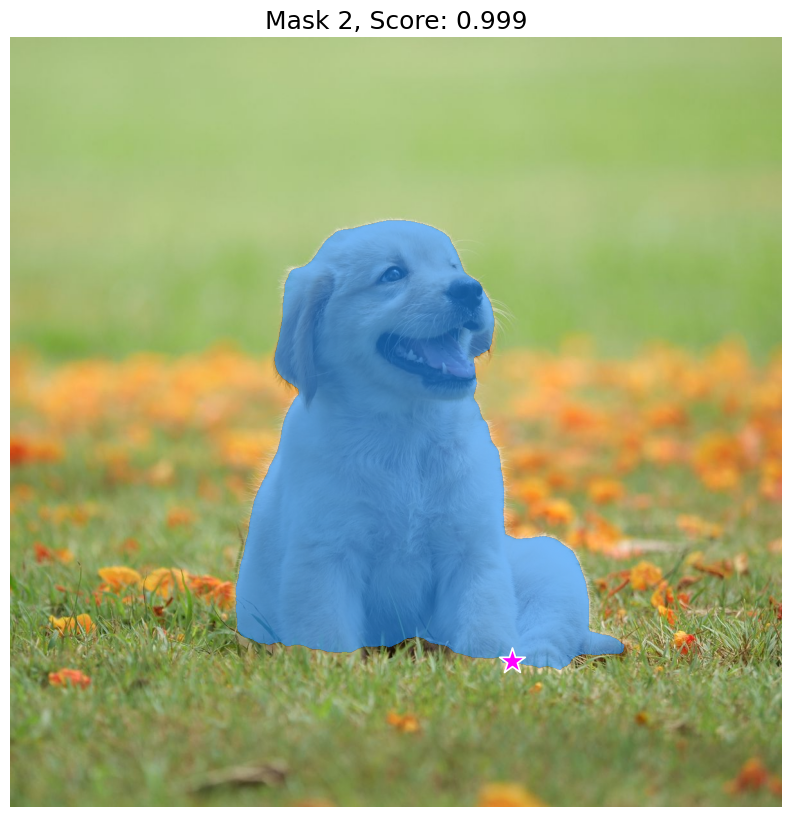} &
    \includegraphics[width=0.185\linewidth , trim=0.1cm 0.1cm 0.1cm 0.55cm, clip]{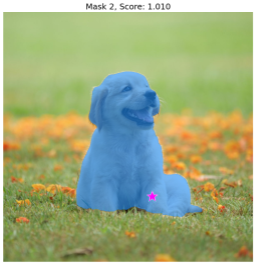}
    \\
    \hline
    \end{tabular}
    \label{tab: SHAP hyperparameters}
    \vspace{-0.78cm}
\end{table}

As previously stated, the explainer has two parameters, with the maximum evaluation parameter being the most important. This parameter determines the region of interest (ROI) and the coordinate with the highest SHAP value (\(P_{\text{coordinates}}\)). The number of evaluation parameters directly impacts the model, as shown in Table~\ref{tab: SHAP hyperparameters}. If the number of parameters is not optimal (such as 100 and 200), the explainer fails to accurately identify the region of interest and instead includes irrelevant areas outside the object. As a result, the SAM segmentation includes other objects backgrounds that are not useful. However, with the number of parameters set to 300, 500, and 1000 the SHAP explainer is able to focus on the specific coordinates that correspond to the object of interest (refer Table.~\ref{tab: SHAP hyperparameters}). It is important to note that increasing the number of evaluation parameters directly increases the computational requirements. Therefore, there needs to be a trade-off between the number of evaluation parameters and the available computational resources. 

\vspace{-0.65cm}
\subsubsection{Image-to-text explanation:}
\label{Subsubsec: results - ImageNet image-to-text explanation}
The SAM's segmented image \(X_v\) serves as the input for the VLM as explained in Section~\ref{Subsec: methodology - image-to-text explanation}. The aforesaid VLMs requires two inputs: the image and the prompt. We design prompts (VLMs instruction) utilising the BLEU scores, as indicated in Table~\ref{tab: prompts for ImageNet}. Although the VLM models were trained to classify,  they do not accurately classify images or identify objects in the images. From Table~\ref{tab: prompts for ImageNet}, it is evident that prompts without actual class labels have low BLEU scores, while prompts with the actual label have high BLEU scores, indicating that the image classifier prediction directly influences the VLMs prediction. Hence, We select the rule-based prompt \(X_{\text{prompt}}\) to explain the prediction. \textbf{\texttt{``Explain the object shown in the image: `\textbf{predicted label}'?''}} The predicted label will be replaced with the actual label predicted by the image classifier. From Table~\ref{tab: prompts for ImageNet}, it is also evident that LLaVA outperforms all other models, hence LLaVA is chosen as the defacto model for further study.

\begin{table}[h!]
    \vspace{-0.7cm}
    \centering
    \caption{Reference explanation for bald eagle from ImageNet dataset.}
    {\fontsize{6}{8}\selectfont
    \vspace{-0.3cm}
    \begin{tabular}{@{}p{0.99\textwidth}@{}}
    \toprule
    \textbf{Reference:} \texttt{This image captures a bald eagle with its wings spread wide. The eagle's body is predominantly brown, with yellow talons and a white head and tail. It is noticeable that the talons appear to be stationed somewhere, and its tail feathers are a lighter shade of brown, adding sense to the image. The eagle's head is turned to the left, and its eyes are focused on something in the distance, so the eagle may be looking for prey or scouting its area.} \\
    \bottomrule
    \end{tabular}
    }
    \label{tab:reference for ImageNet}
    \vspace{-0.7cm}
\end{table}

\begin{table}[t!]
    \vspace{-0.35cm}
    \centering
    \caption{BLEU scores for different prompts for bald eagle.}
    \vspace{-0.25cm}
    \resizebox{\linewidth}{!}
    {\fontsize{6}{8}\selectfont
    \begin{tabular}{|l|c|c|c|c|c|}
    \hline
    \multirow{2}{*}{\textbf{Prompt}} & \multicolumn{5}{c|}{\textbf{BLEU Score}} \\ 
    \cline{2-6}
    & \textbf{LLaVA} & \textbf{Instruct-BLIP} & \textbf{GIT} & \textbf{MiniCPM} & \textbf{InternLM} \\ 
    \hline
    \texttt{Explain?} & 0.1171 & 0.0000 & 0.0000 & 0.0958 & 0.0514 \\
    \texttt{Explain the object in the image?} & 0.1627 & 0.0415 & 0.0000 & 0.1031 & 0.0673 \\
    \texttt{Explain the object in the image: `eagle'?} & 0.2627 & 0.0000 & 0.0000 & 0.0000 & 0.0724 \\
    \texttt{Explain the object in the image: `bird'?} & 0.1293 & 0.0000 & 0.0000 & 0.0632 & 0.0629 \\
    \texttt{Explain the object in the image: `bald\_eagle'?} & 0.2668 & 0.0000 & 0.0000 & 0.0177 & 0.0652 \\ 
    \hline
    \end{tabular}
    }
    \label{tab: prompts for ImageNet}
    \vspace{-0.75cm}
\end{table}

\begin{figure}[!h]
    \vspace{-0.8cm}
    \centering
    \includegraphics[width=0.9\textwidth]{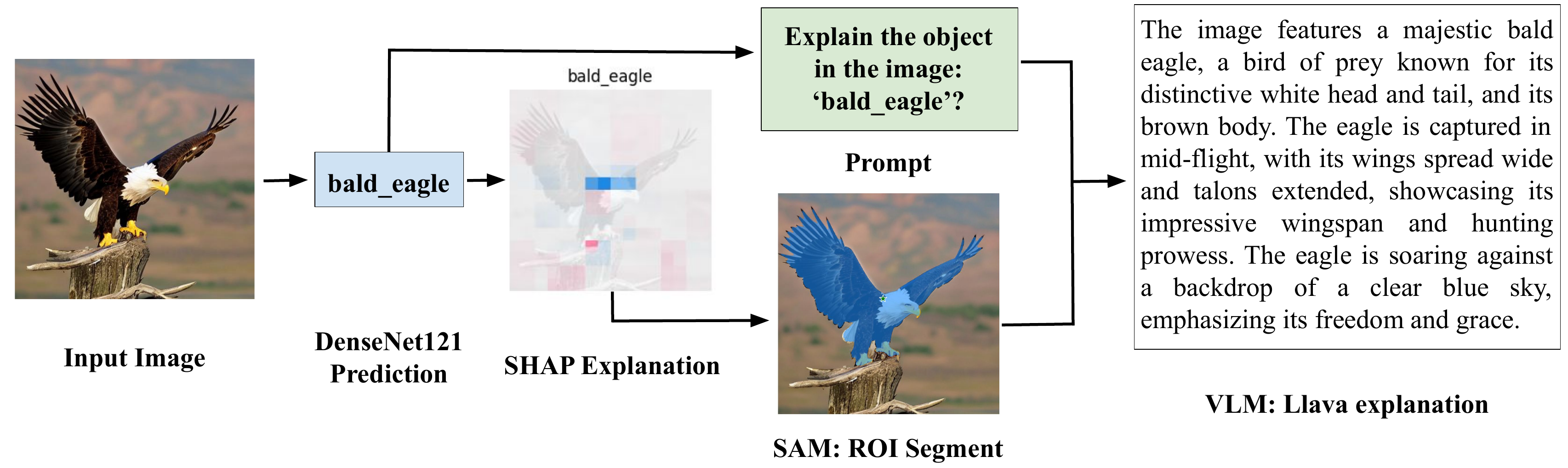}
    \vspace{-0.3cm}
    \caption{Explanation for Bald Eagle image illustrating the pipeline.}
    \label{fig: bald eagle :output flow w.r.t architecture}
    \vspace{-0.75cm}
\end{figure}

The output of the captioning model for the bald eagle is depicted in Fig~\ref{fig: bald eagle :output flow w.r.t architecture}. The descriptive explanation for the bald eagle has mentioned the features of the bird: a white head and tail and its brown body. It also mentioned mid-flight, with its wings spread wide and talons extended. The explanation of the image specifically mid-flight represents only the segmented input, ignoring the background, because birds usually spread their wings on flight. These explicit and detailed explanations, combined with visual representations, offer a more concrete and tangible explanation.

\vspace{-0.55cm}
\subsection{VALE for SONAR Image Classifier}
\label{Subsec:  on sonar image classication}
\vspace{-0.25cm}
In this section, we assess the efficacy of the proposed VALE architecture  with a custom dataset and custom-built classification model for an `in-the-wild' deployment scenario. As explained in Section~\ref{Subsec: methodology - image classification}, transfer learning with the backbone model DenseNet121~\cite{huang2017densely} is used to develop a tri-class image classification model on the SONAR dataset. Since the dataset is imbalanced, we use synthetic image generation techniques such as flipping, cropping, and rotation to balance it~\cite{rawat2017deep}. We also apply a stratified random sampling approach to split the dataset into train, validate, and test sets, which allows us to effectively assess the image classifier's performance. Following extensive training, the classifier produces a training accuracy of 99.32\% with 14 epochs. Classifier performance on the validation set and test set are 96.33\% and 96\%, respectively. The per-class classification report for the test dataset is given in Table~\ref{tab:classification_report}.

\begin{table}[htph!]
    \vspace{-0.65cm}
    \begin{center}
    \caption{Classification Report for the SONAR Image Classifier.}
    \vspace{-0.35cm}
    \begin{tabular}{|@{}l c@{\hspace{0.5cm}} c@{\hspace{0.5cm}} c@{\hspace{0.5cm}} c@{}|}
    \hline
    \textbf{Class} & \textbf{Accuracy} & \textbf{Precision} & \textbf{Recall} & \textbf{F1-Score} \\
    \hline
    Airplane & 0.96 & 0.90 & 0.99 & 0.94  \\
    Seafloor & 0.98 & 0.99 & 1.00 & 0.99  \\
    Ship & 0.92 & 1.00 & 0.89 & 0.94   \\
    \hline
    \end{tabular}
    \label{tab:classification_report}
    \end{center}
    \vspace{-0.7cm}
\end{table}

\vspace{-0.55cm}
\subsubsection{Proposed Framework with SONAR Image classifier:}
\label{Subsubsec: results - SONAR image classifier and  explainer}
\vspace{-0.5cm}

\begin{center}
\begin{table}[t!]
    \centering
    \vspace{-0.35cm}
    \caption{Visual Explanations for SONAR dataset}
    \vspace{-0.30cm}
    {\fontsize{6}{8}\selectfont
    \begin{tabular}{|c|c|c|c|}
    \hline
    \textbf{Original Image} & \textbf{SHAP Explanation} & \textbf{Region of Interest} & \textbf{Mask} \\
    \hline
    \includegraphics[width=0.18\linewidth, trim=1.2cm 0.90cm 0.1cm 0.1cm, clip]{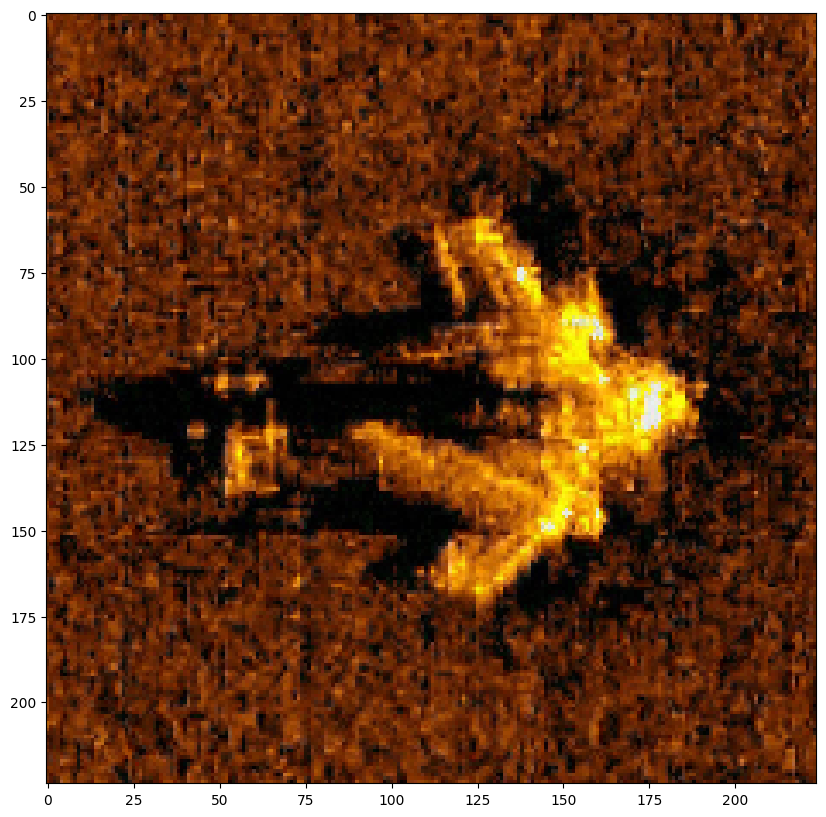} & 
    \includegraphics[width=0.18\linewidth]{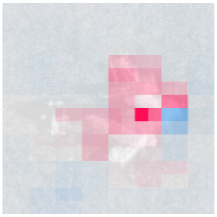} & 
    \includegraphics[width=0.18\linewidth, trim=1.2cm 0.90cm 0.1cm 0.1cm, clip]{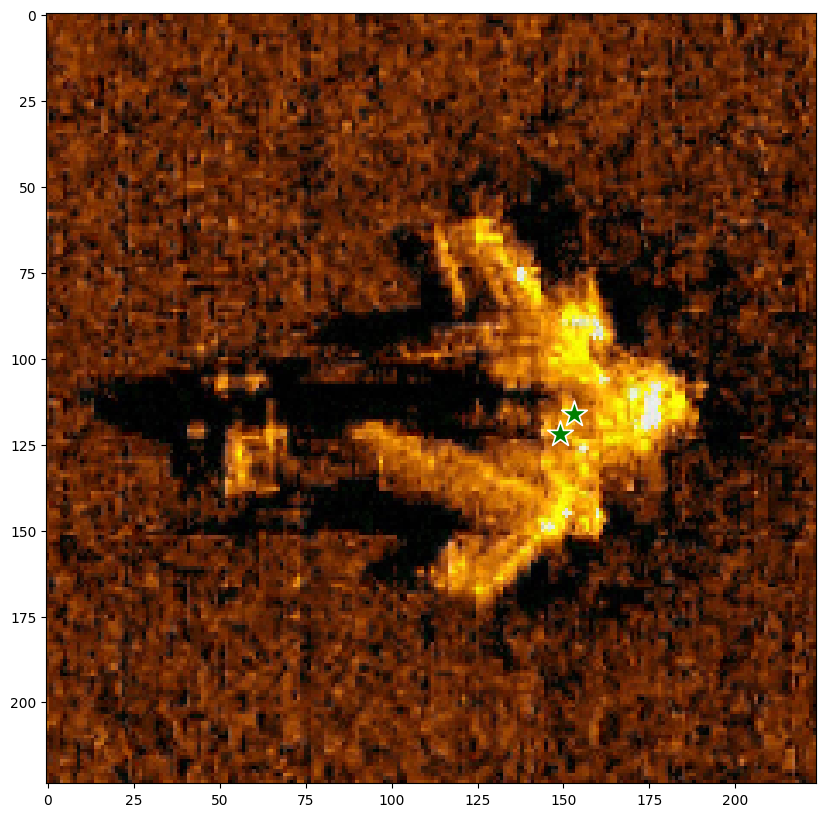} & \includegraphics[width=0.18\linewidth,  trim=0.1cm 0.1cm 0.1cm 0.87cm, clip]{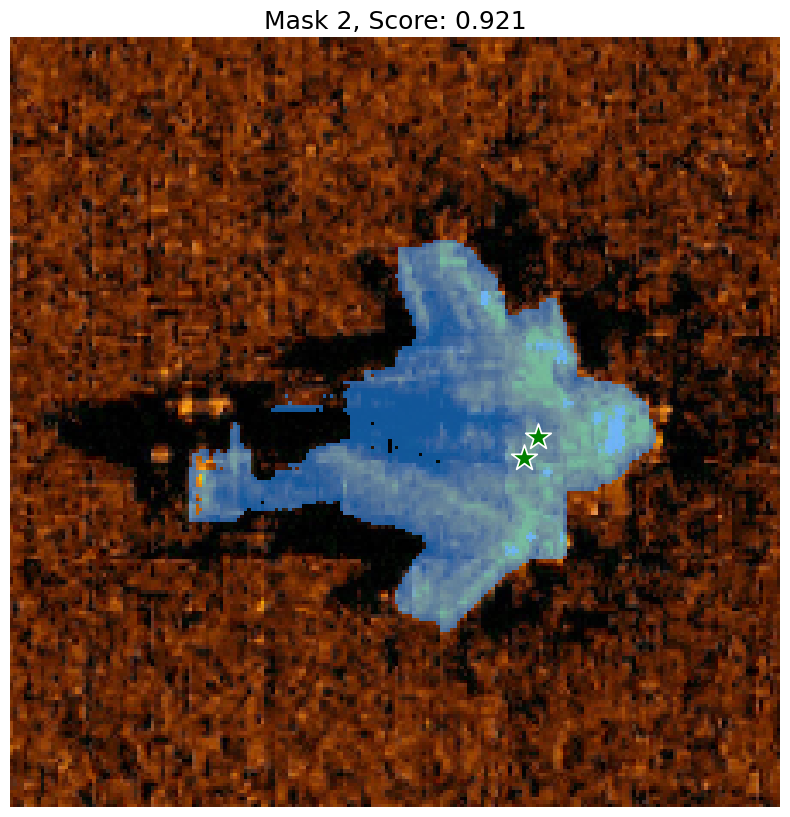} \\
    \hline
    \includegraphics[width=0.18\linewidth, trim=1.2cm 0.90cm 0.1cm 0.1cm, clip]{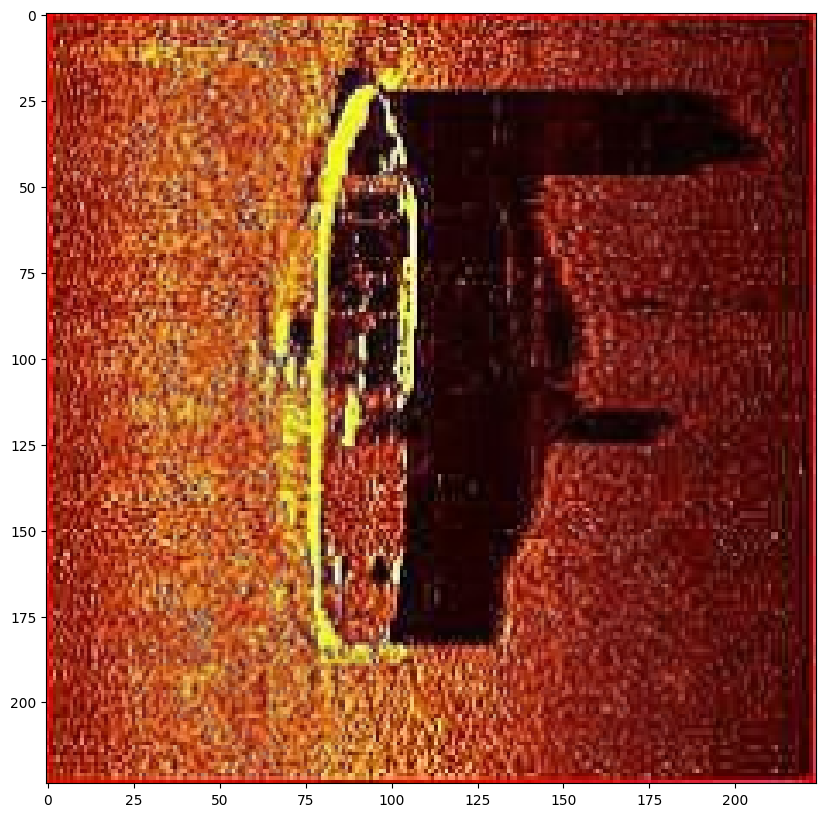} & 
    \includegraphics[width=0.18\linewidth]{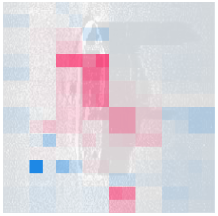} & 
    \includegraphics[width=0.18\linewidth, trim=1.2cm 0.90cm 0.1cm 0.1cm, clip]{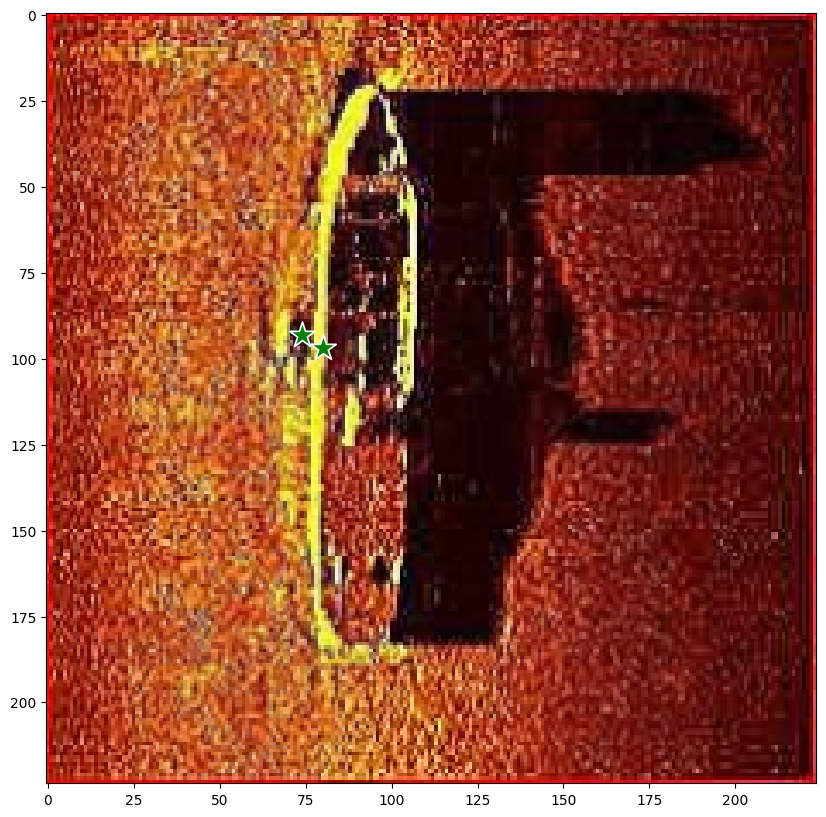} & \includegraphics[width=0.18\linewidth,  trim=0.1cm 0.1cm 0.1cm 0.87cm, clip]{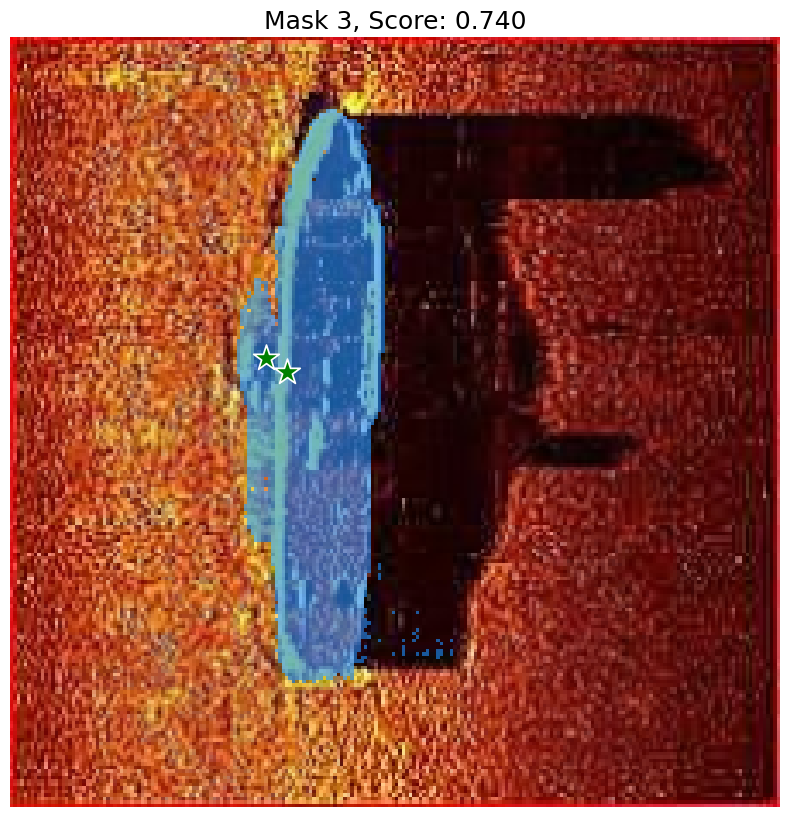} \\
    \hline
    \end{tabular}
    }
    \label{tab:sonar analysis results}
    \vspace{-0.75cm}
\end{table}
\end{center}

\vspace{-0.95cm}
The trained sonar image classifier predicted the image in first row of Table~\ref{tab: explanation samples} as "Airplane". Using the proposed architecture , this prediction is further explained using SHAP explanations. Instead of one prompt for SAM, we utilize the top two (\(P_{\text{coordinates}}\)). Note that the number of prompts does not increase the computation but it helps in better segmentation, especially in low-quality or pixelated images as depicted in Table~\ref{tab:sonar analysis results}. The prompt `\textbf{\texttt{Explain the object in the image: ‘Airplane’?}}' along with the segmented image provides the textual description as depicted in Table~\ref{tab: explanation samples}. Although the description appears satisfactory, it can be further refined to offer an explanation even for images of extremely poor quality images by tuning the prompt.

\vspace{-0.55cm}
\subsubsection{Prompt Engineering to fine-tune VALE:}
\label{Subsubsec: results - SONAR prompt engineering}
The explainer has the ability to provide explanations for predictions using a default prompt input. However, its effectiveness is limited because the VLM is trained on a general dataset, which means that it cannot provide specific explanations without prompt engineering.

\begin{table}[h!]
    \vspace{-0.6cm}
    \centering
    \caption{Reference explanations for SONAR Dataset}
    \vspace{-0.35cm}
    {\fontsize{6}{8}\selectfont
    \begin{tabular}{@{}p{0.99\textwidth}@{}}
    \midrule
    \texttt{Plane: The image features a pixelated representation of an object that resembles an airplane captured using SONAR at a grazing angle of 70 to 80 degrees because the shadow is on the right side of the image. The wings and the body are clearly visible; however, the part of the wings is buried in the sand, which shows that it could be seafloor, and by the shape and body of the plane, we could conclude that it could be a war plane. I can also notice the broken tails, which represents that there could have been an accident.} \\
    \midrule
    \texttt{Seafloor: The image represents the seafloor, which appears to be textured, rough, and uneven, as is typical of seafloors, and contains no objects.} \\
    \midrule
    \texttt{Ship: The image features a pixelated representation of a ship captured using SONAR at a grazing angle of 75 to 85 degrees because the left side of the ship is visible but the right side is not. The curved head, tail, and body are clearly visible; I can also notice container-like shapes, and the hull is visible with its windows and portholes aligned along the sides, indicative of a yacht design or cargo carrier that accommodates living quarters inside.} \\
    \bottomrule
    \end{tabular}
    \label{tab:sonar_image_descriptions}
    \vspace{-0.85cm}
    }
\end{table}

\begin{table}[h]
    \vspace{-0.6cm}
    \centering
    \caption{BLEU Scores for Different Prompts on the SONAR dataset.}
    \vspace{-0.35cm}
    {\fontsize{6}{8}\selectfont
    \begin{tabularx}{\linewidth}{|@{\centering\arraybackslash}p{7cm}>{\centering\arraybackslash}p{1.61cm}>{\centering\arraybackslash}p{1.6cm}>{\centering\arraybackslash}p{1.615cm}|}
    \hline 
    \textbf{Prompt} & \textbf{Airplane} & \textbf{Seafloor} & \textbf{Ship} \\
    \hline 
    \texttt{Explain the object in the image: \{predicted label\}?} & 0.0732 & 0.0000 & 0.0567 \\
    \texttt{Describe only the object in the image that represents the \{predicted label\} as acquired through the use of synthetic aperture sonar, make sure to ignore the background?} & 0.1216 & 0.1954 & 0.2695 \\
    \hline 
    \end{tabularx}
    }
    \label{tab:bleu scores for SONAR}
    \vspace{-.75cm}
\end{table}

\begin{table}[h!]
\vspace{-0.35cm}
    \centering
    \caption{Explanation samples: First column: original image; second column: segmented image; third column: prompt 1; fourth column: Prompt 2}
    \vspace{-0.35cm}
    {\fontsize{6}{8}\selectfont
    \begin{tabularx}{\linewidth}{|p{1.45cm}|p{1.45cm}|p{4.42cm}|p{4.42cm}|}
        \hline
        \textbf{Original \newline Image} & \textbf{Segmented Image} & \textbf{Explain the object in the image: {predicted label}?} & \textbf{Describe only the object in the image that represents the {predicted label} as acquired through the use of synthetic aperture sonar, make sure to ignore the background?} \\ 
        \hline
        \includegraphics[width=1.45cm,height=1.45cm, trim=0.88cm 0.7cm 0.1cm 0.1cm, clip]{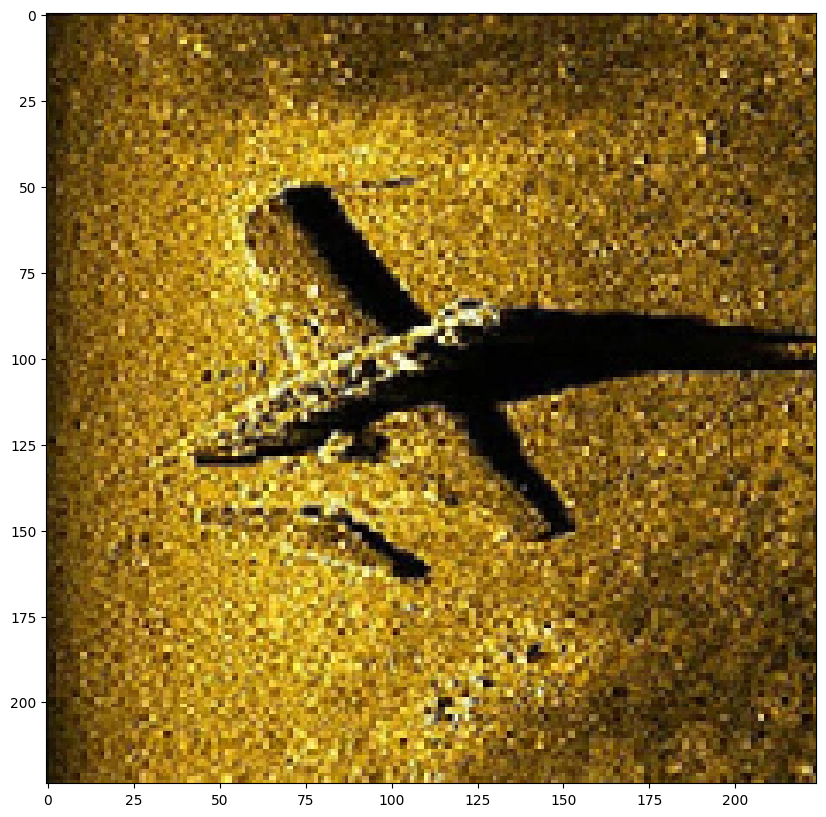} & \includegraphics[width=1.45cm,height=1.45cm, trim=0.1cm 0.1cm 0.1cm 0.87cm, clip]{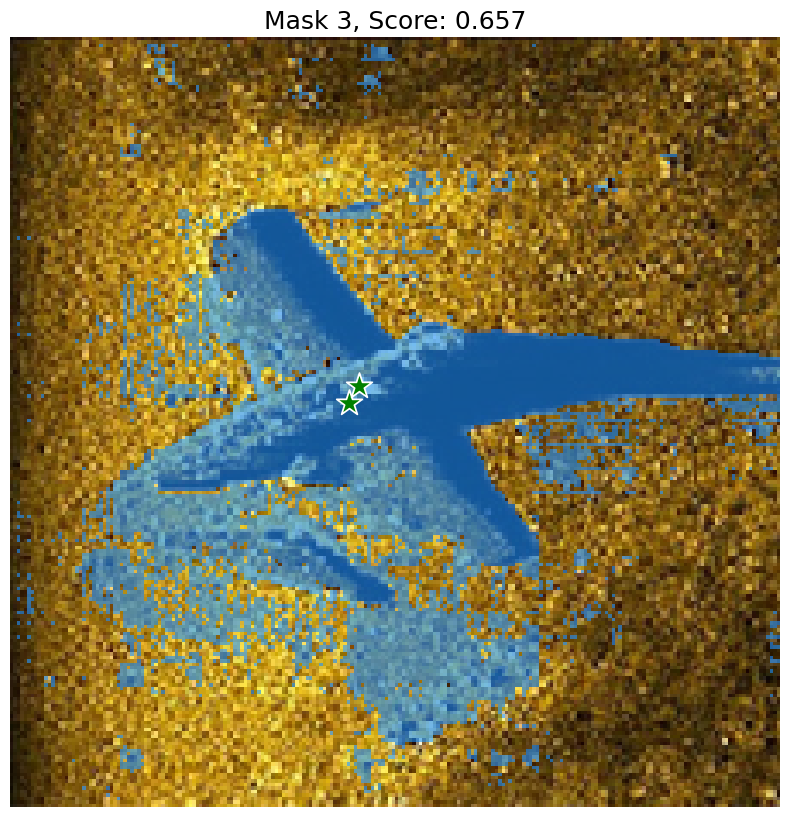} & \vspace{-1.4cm}The image features a close-up view of a small airplane, which appears to be made of wood. The airplane is positioned diagonally, with its nose pointing towards the top left corner of the image and its tail towards the bottom right corner. The airplane's design is intricate, with a pointed nose and a tail fin, giving it a unique and detailed appearance. The image does not contain any text or other discernible objects. The airplane's position and the absence of other objects suggest that it is the main focus of the image. & \vspace{-1.4cm} The image you've provided appears to be a low-resolution or pixelated representation of an object that resembles an airplane. Due to the low resolution and the pixelation, it's difficult to provide specific details about the object. However, it seems to have a general shape that could be interpreted as an airplane, with a body, wings, and possibly a tail section. The image quality does not allow for a detailed description of the object's features or condition. \\
        \hline
        \includegraphics[width=1.45cm,height=1.45cm, trim=0.88cm 0.7cm 0.1cm 0.1cm, clip]{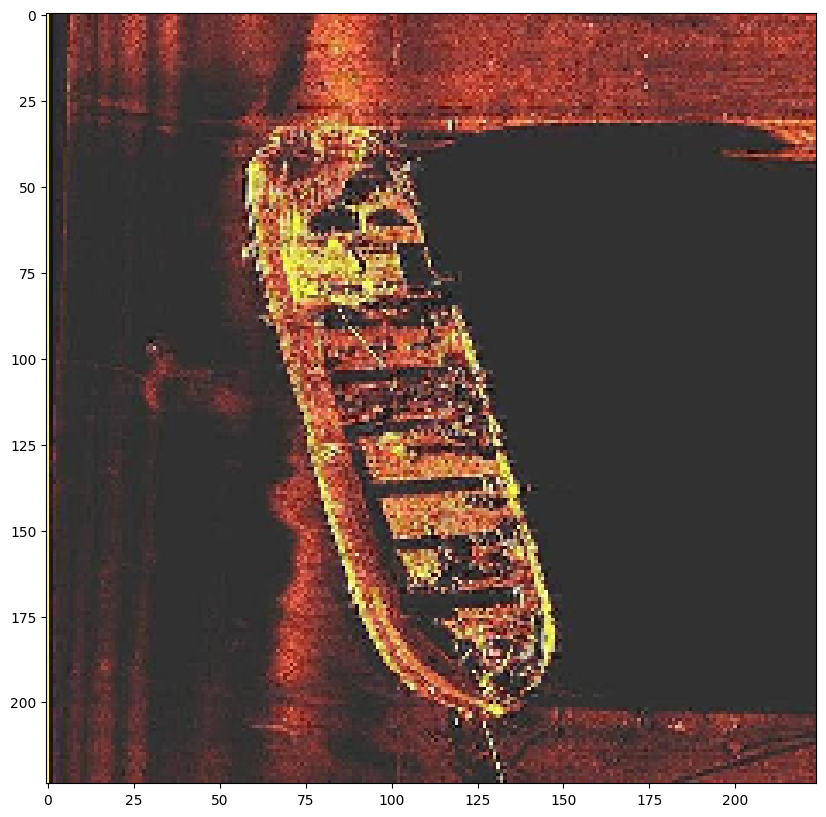} & \includegraphics[width=1.45cm,height=1.45cm, trim=0.1cm 0.1cm 0.1cm 0.87cm, clip]{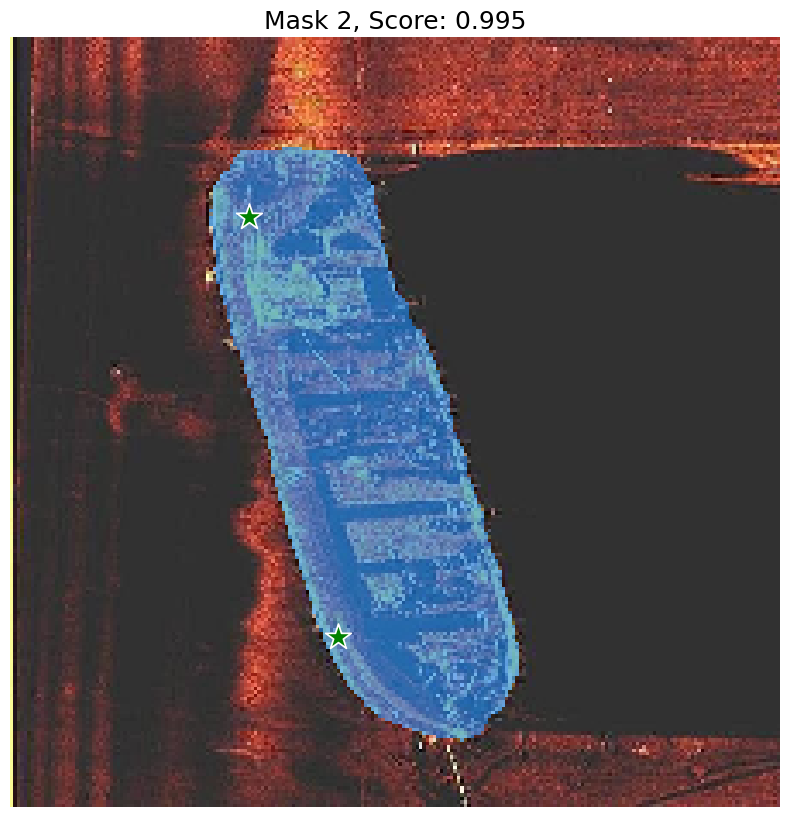} & \vspace{-1.4cm}The image features a wooden object that appears to be a ship or a boat. The object is brown in color and has a curved shape, reminiscent of a boat or a ship. The pixelated nature makes it difficult to determine the specific details and what the object is. Based on the limited information given by this image, it might be defining the shape of a ship, but without higher resolution or additional context, I'm unable to provide a precise description or identification & \vspace{-1.4cm} The image you've provided appears to show a yacht. This type of image typically represents a ship captured using synthetic aperture sonar, an underwater mapping technique that uses sonar equipment to generate high-resolution images of the seabed and submerged objects. The yacht's hull is visible with its windows and portholes aligned along the sides, indicative of a yacht design that accommodates living quarters inside. The image quality suggests that the synthetic aperture sonar has successfully captured the characteristics of the ship's underwater structure, which is useful for maritime archaeology, survey work, and underwater exploration. \\
        \hline
    \end{tabularx}
    }
    \label{tab: explanation samples}
\vspace{-0.65cm}
\end{table}

The BLEU score and its corresponding description obtained from two different prompts with reference explanations from Table~\ref{tab:sonar_image_descriptions} are provided in Table~\ref{tab:bleu scores for SONAR} and Table~\ref{tab: explanation samples}, respectively.For the plane image, the default prompt mentions a wooden plane with a BLEU score of 0.0732, while the customized prompt `\textbf{\texttt{Describe only the object in the image that represents the `Airplane' as \newline acquired through the use of synthetic aperture sonar, make sure to \newline ignore the background?}}' provides a better explanation mentioning the features of the plane and the quality of the captured image, resulting in an improved score of 0.1216. Similarly, for the ship image in Table~\ref{tab: explanation samples}, the default prompt gave a satisfactory explanation, mentioning its color and structure, as well as incorrectly identified it as a wooden boat, a toy with a score of 0.0567. On the other hand, the customized prompt provides a superior explanation, correctly identifying it as a yacht with visible windows and portholes aligned along the sides, achieving a BLEU score of 0.2695. Therefore, for a custom dataset, an optimal prompt should be engineered to provide an accurate explanation.

\vspace{-0.5cm}
\subsection{State-of-the-art Comparison}
\label{Subsec: SOTA}
\vspace{-0.25cm}
To showcase the efficacy of the VALE architecture, both qualitative and quantitative analysis is conducted with state-of-the-art approaches. Our work represents the very first attempt of a multimodal explainer utilizing XAI. As a result, there are no comparative metrics available. However, we have provided a summary of similar XAI approaches in Table~\ref{tab: SOTA comparison}. Sahay et al.~\cite{sahay2021approach} employed LIME to generate textual explanations. Bennetot et al.~\cite{bennetot2022greybox} utilized an encoder-decoder architecture to offer textual explanations for the corresponding visual counterpart. Another recent work Sun et al.~\cite{sun2024explain} used LIME + SHAP to provide visual explanations using SAM. To the best of our knowledge, there is no existing work that offers textual explanations from the visual counterpart obtained from an explainer. In this work, we employed SHAP and pre-trained models to provide comprehensive explanations using both visual and textual components.

\begin{table}[h]
    \vspace{-0.5cm}
    \centering
    \caption{State-of-the-art Comparison}
    \vspace{-0.25cm}
    \begin{tabularx}{\linewidth}{X X X}
        \hline
        \textbf{Approach} & \textbf{Explainer} & \textbf{Mode of Explanation} \\
        \hline
        Sahay et al.~\cite{sahay2021approach} & LIME & Visual Explanation \\
        Bennetot et al.~\cite{bennetot2022greybox} & Encoder-Decoder & Textual Explanation \\
        Sun et al.~\cite{sun2024explain} & LIME + SHAP & Visual Explanation \\
         (Ours) & SHAP & Both Visual and Textual Explanation \\
        \hline
        \hline
    \end{tabularx}
    \label{tab: SOTA comparison}
    \vspace{-0.55cm}
\end{table}

\vspace{-.65cm}
\section{Conclusion and Future Works}
\label{Sec: Conclusion}
\vspace{-.35cm}
This work presents a novel \textbf{multimodal Visual and Language Explanation framework (VALE)} based on a explainer for the first time in the XAI paradigm to explain the predictions made by image classifiers. The efficacy of VALE on the general ImageNet dataset and the specific underwater SONAR datasets is demonstrated. In both the cases, VALE highlighted the superior performance by integrating the SAM and VLM models within the XAI framework that reduces the semantic gap and boosts interpretability and confidence. The use-case scenario for classifying underwater objects using SONAR imagery further highlighted the practicality of in the wild. Future research aims to improve explainer efficacy by integrating additional XAI techniques like LIME and LRP.

\vspace{-0.525cm}

\section{Acknowledgements}
\label{sec: acknowledgements}
\vspace{-0.35cm}
This work was partially supported by the Naval Research Board (NRB), DRDO, Government of India under grant number: NRB/505/SG/22-23.

\bibliographystyle{splncs04}
\bibliography{VALE_Explainer}

\begin{thebibliography}{10}
\providecommand{\url}[1]{\texttt{#1}}
\providecommand{\urlprefix}{URL }
\providecommand{\doi}[1]{https://doi.org/#1}

\bibitem{abdollahi2021urban}
Abdollahi, A., Pradhan, B.: Urban vegetation mapping from aerial imagery using explainable ai (xai). Sensors  \textbf{21}(14), ~4738 (2021)

\bibitem{ayush2020generating}
Ayush, K., Uzkent, B., Burke, M., Lobell, D., Ermon, S.: Generating interpretable poverty maps using object detection in satellite images. arXiv preprint arXiv:2002.01612  (2020)

\bibitem{bach2015pixel}
Bach, S., Binder, A., Montavon, G., Klauschen, F., M{\"u}ller, K.R., Samek, W.: On pixel-wise explanations for non-linear classifier decisions by layer-wise relevance propagation. PloS one  \textbf{10}(7),  e0130140 (2015)

\bibitem{bennetot2022greybox}
Bennetot, A., Franchi, G., Del~Ser, J., Chatila, R., Diaz-Rodriguez, N.: Greybox xai: A neural-symbolic learning framework to produce interpretable predictions for image classification. Knowledge-Based Systems  \textbf{258},  109947 (2022)

\bibitem{chungath2023transfer}
Chungath, T.T., Nambiar, A.M., Mittal, A.: Transfer learning and few-shot learning based deep neural network models for underwater sonar image classification with a few samples. IEEE Journal of Oceanic Engineering  (2023)

\bibitem{dai2023instructblip}
Dai, W., Li, J., Li, D., Tiong, A., Zhao, J., Wang, W., Li, B., Fung, P., Hoi, S.: Instruct{BLIP}: Towards general-purpose vision-language models with instruction tuning. In: Conference on Neural Information Processing Systems (2023)

\bibitem{deng2009imagenet}
Deng, J., Dong, W., Socher, R., Li, L.J., Li, K., Fei-Fei, L.: Imagenet: A large-scale hierarchical image database. In: 2009 IEEE conference on computer vision and pattern recognition. pp. 248--255. Ieee (2009)

\bibitem{dewi2023xai}
Dewi, C., CHEN, R.C., Yu, H., JIANG, X.: Xai for image captioning using shap. Journal of Information Science \& Engineering  \textbf{39}(4) (2023)

\bibitem{divvala2009empirical}
Divvala, S.K., Hoiem, D., Hays, J.H., Efros, A.A., Hebert, M.: An empirical study of context in object detection. In: 2009 IEEE Conference on computer vision and Pattern Recognition. pp. 1271--1278. IEEE (2009)

\bibitem{dong2024internlm}
Dong, X., Zhang, P., Zang, Y., Cao, Y., Wang, B., Ouyang, L., Wei, X., Zhang, S., Duan, H., Cao, M., et~al.: Internlm-xcomposer2: Mastering free-form text-image composition and comprehension in vision-language large model. arXiv preprint arXiv:2401.16420  (2024)

\bibitem{han2020explainable}
Han, S.H., Kwon, M.S., Choi, H.J.: Explainable ai (xai) approach to image captioning. The Journal of Engineering  \textbf{2020}(13),  589--594 (2020)

\bibitem{hu2024minicpm}
Hu, S., Tu, Y., Han, X., He, C., Cui, G., Long, X., Zheng, Z., Fang, Y., Huang, Y., Zhao, W., et~al.: Minicpm: Unveiling the potential of small language models with scalable training strategies. arXiv preprint arXiv:2404.06395  (2024)

\bibitem{huang2017densely}
Huang, G., Liu, Z., Van Der~Maaten, L., Weinberger, K.Q.: Densely connected convolutional networks. In: Proceedings of the IEEE conference on computer vision and pattern recognition. pp. 4700--4708 (2017)

\bibitem{huo2020underwater}
Huo, G., Wu, Z., Li, J.: Underwater object classification in sidescan sonar images using deep transfer learning and semisynthetic training data. IEEE access  (2020)

\bibitem{kirillov2023segment}
Kirillov, A., Mintun, E., Ravi, N., Mao, H., Rolland, C., Gustafson, L., Xiao, T., Whitehead, S., Berg, A.C., Lo, W.Y., et~al.: Segment anything. In: Proceedings of the IEEE/CVF International Conference on Computer Vision (2023)

\bibitem{liu2024visual}
Liu, H., Li, C., Wu, Q., Lee, Y.J.: Visual instruction tuning. Advances in neural information processing systems  \textbf{36} (2024)

\bibitem{lundberg2017unified}
Lundberg, S.M., Lee, S.I.: A unified approach to interpreting model predictions. Advances in neural information processing systems  \textbf{30} (2017)

\bibitem{o2015introduction}
O'shea, K., Nash, R.: An introduction to convolutional neural networks. arXiv preprint arXiv:1511.08458  (2015)

\bibitem{papineni2002bleu}
Papineni, K., Roukos, S., Ward, T., Zhu, W.J.: Bleu: a method for automatic evaluation of machine translation. In: Proceedings of the 40th annual meeting of the Association for Computational Linguistics. pp. 311--318 (2002)

\bibitem{rawat2017deep}
Rawat, W., Wang, Z.: Deep convolutional neural networks for image classification: A comprehensive review. Neural computation  \textbf{29}(9),  2352--2449 (2017)

\bibitem{ribeiro2016should}
Ribeiro, M.T., Singh, S., Guestrin, C.: " why should i trust you?" explaining the predictions of any classifier. In: Proceedings of the 22nd ACM SIGKDD international conference on knowledge discovery and data mining. pp. 1135--1144 (2016)

\bibitem{sahay2021approach}
Sahay, S., Omare, N., Shukla, K.: An approach to identify captioning keywords in an image using lime. In: 2021 International Conference on Computing, Communication, and Intelligent Systems (ICCCIS). pp. 648--651. IEEE (2021)

\bibitem{samek2017explainable}
Samek, W., Wiegand, T., M{\"u}ller, K.R.: Explainable artificial intelligence: Understanding, visualizing and interpreting deep learning models. arXiv preprint arXiv:1708.08296  (2017)

\bibitem{selvaraju2017grad}
Selvaraju, R.R., Cogswell, M., Das, A., Vedantam, R., Parikh, D., Batra, D.: Grad-cam: Visual explanations from deep networks via gradient-based localization. In: Proceedings of the IEEE international conference on computer vision (2017)

\bibitem{sun2024explain}
Sun, A., Ma, P., Yuan, Y., Wang, S.: Explain any concept: Segment anything meets concept-based explanation. Advances in Neural Information Processing  (2024)

\bibitem{vujovic2021classification}
Vujovi{\'c}, {\v{Z}}., et~al.: Classification model evaluation metrics. International Journal of Advanced Computer Science and Applications  \textbf{12}(6),  599--606 (2021)

\bibitem{wang2022git}
Wang, J., Yang, Z., Hu, X., Li, L., Lin, K., Gan, Z., Liu, Z., Liu, C., Wang, L.: Git: A generative image-to-text transformer for vision and language. arXiv preprint arXiv:2205.14100  (2022)

\bibitem{winter2002shapley}
Winter, E.: The shapley value. Handbook of game theory with economic applications  \textbf{3},  2025--2054 (2002)

\bibitem{xu2015show}
Xu, K., Ba, J., Kiros, R., Cho, K., Courville, A., Salakhudinov, R., Zemel, R., Bengio, Y.: Show, attend and tell: Neural image caption generation with visual attention. In: International conference on machine learning (2015)

\bibitem{zhang2021self}
Zhang, P., Tang, J., Zhong, H., Ning, M., Liu, D., Wu, K.: Self-trained target detection of radar and sonar images using automatic deep learning. IEEE Transactions on Geoscience and Remote Sensing  \textbf{60},  1--14 (2021)

\end{thebibliography}
\vspace{-5cm}
\end{document}